\documentclass[10pt,twocolumn,letterpaper]{article}

\usepackage{cvpr}              

\usepackage{graphicx}
\usepackage{amsmath}
\usepackage{amssymb}
\usepackage{booktabs}
\usepackage{xcolor}
\usepackage{color, colortbl}
\usepackage{verbatim}
\usepackage{csquotes}
\usepackage{xcolor}
\usepackage{rotating}

\usepackage{algorithm}
\usepackage{algorithmic}
\usepackage{tabularx}
\usepackage{multirow}
\usepackage{multicol}
\usepackage{bm}
\usepackage{booktabs, array}
\usepackage{pifont}
\usepackage{amsmath}
\usepackage{gradient-text}
\usepackage{epstopdf}
\usepackage[titletoc]{appendix}
\usepackage[accsupp]{axessibility} 

\definecolor{cvprblue}{rgb}{0.21,0.49,0.74}
\usepackage[pagebackref,breaklinks,colorlinks,allcolors=cvprblue]{hyperref}

\newcommand{\cmark}{\ding{51}}
\newcommand{\xmark}{\ding{55}}

\definecolor{mypurple}{RGB}{60,52,118} 
\definecolor{mygreen}{RGB}{58,193,58}
\definecolor{myred}{RGB}{172,77,108}

\def\paperID{30693} 
\def\confName{CVPR}
\def\confYear{2026}

\title{V2U4Real: A Real-world Large-scale Dataset for Vehicle-to-UAV Cooperative Perception}

\author{
Weijia Li$^{1,2}$\textsuperscript{*},
Haoen Xiang$^{1,2}$\textsuperscript{*},
Tianxu Wang$^{1,2}$,
Shuaibing Wu$^{1,2}$,\\
Qiming Xia$^{1,2}$,
Cheng Wang$^{1,2}$,
Chenglu Wen$^{1,2}$\textsuperscript{$\dagger$}\\
$^{1}$Fujian Key Laboratory of Urban Intelligent Sensing and Computing, Xiamen University, China \\
$^{2}$Key Laboratory of Multimedia Trusted Perception and Efficient Computing,\\
Ministry of Education of China, Xiamen University, China
}

\begin{document}

\twocolumn[{%
    \renewcommand\twocolumn[1][]{#1}%
    \maketitle%
    \begin{center}
        \vspace{-2em}
        \begin{minipage}{0.49\textwidth}
        \vspace{0.3em}
            \centering
            \includegraphics[width=\textwidth]{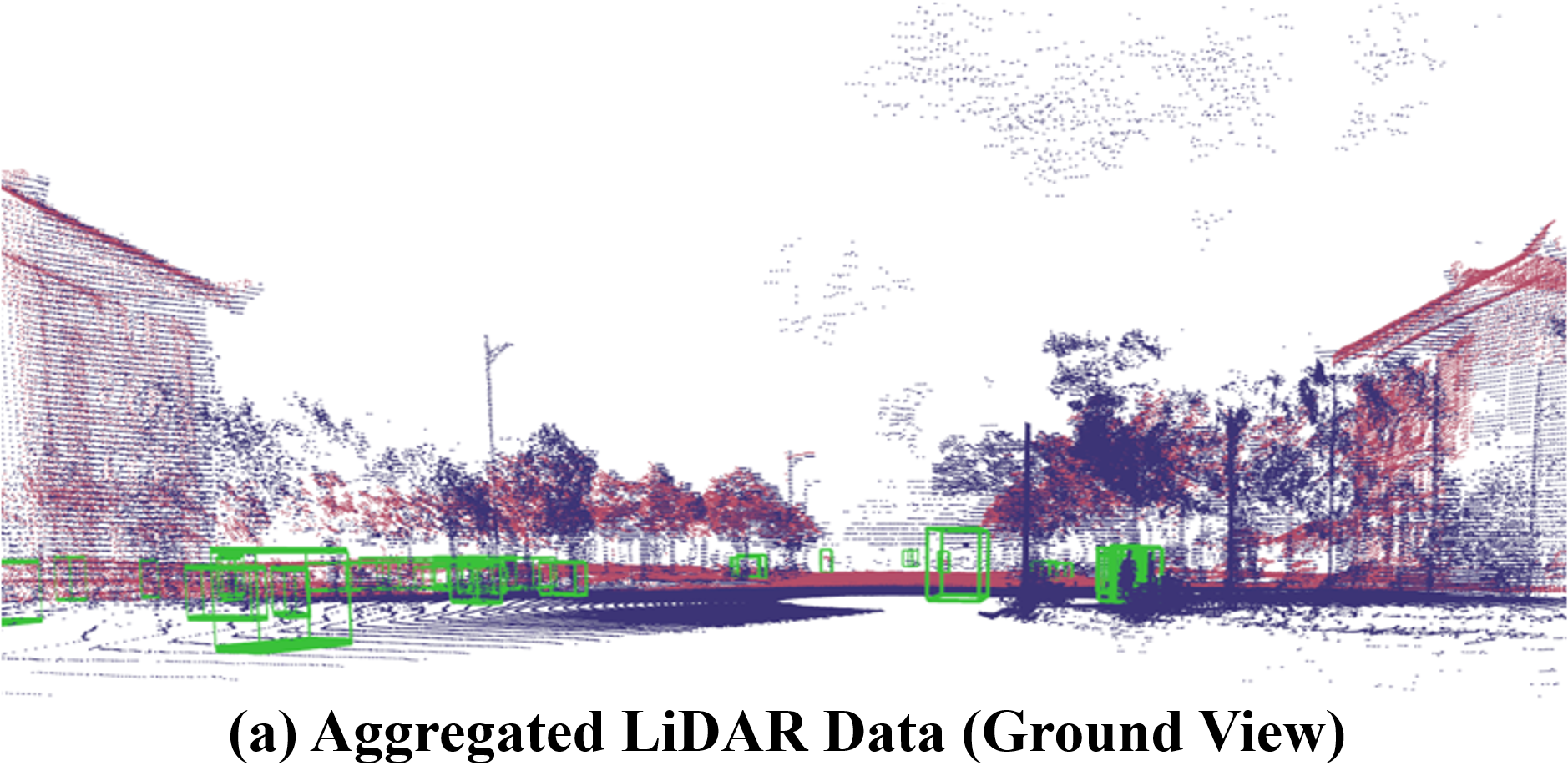}
        \end{minipage}
        \hfil
        \begin{minipage}{0.49\textwidth}
            \centering
            \includegraphics[width=\textwidth]{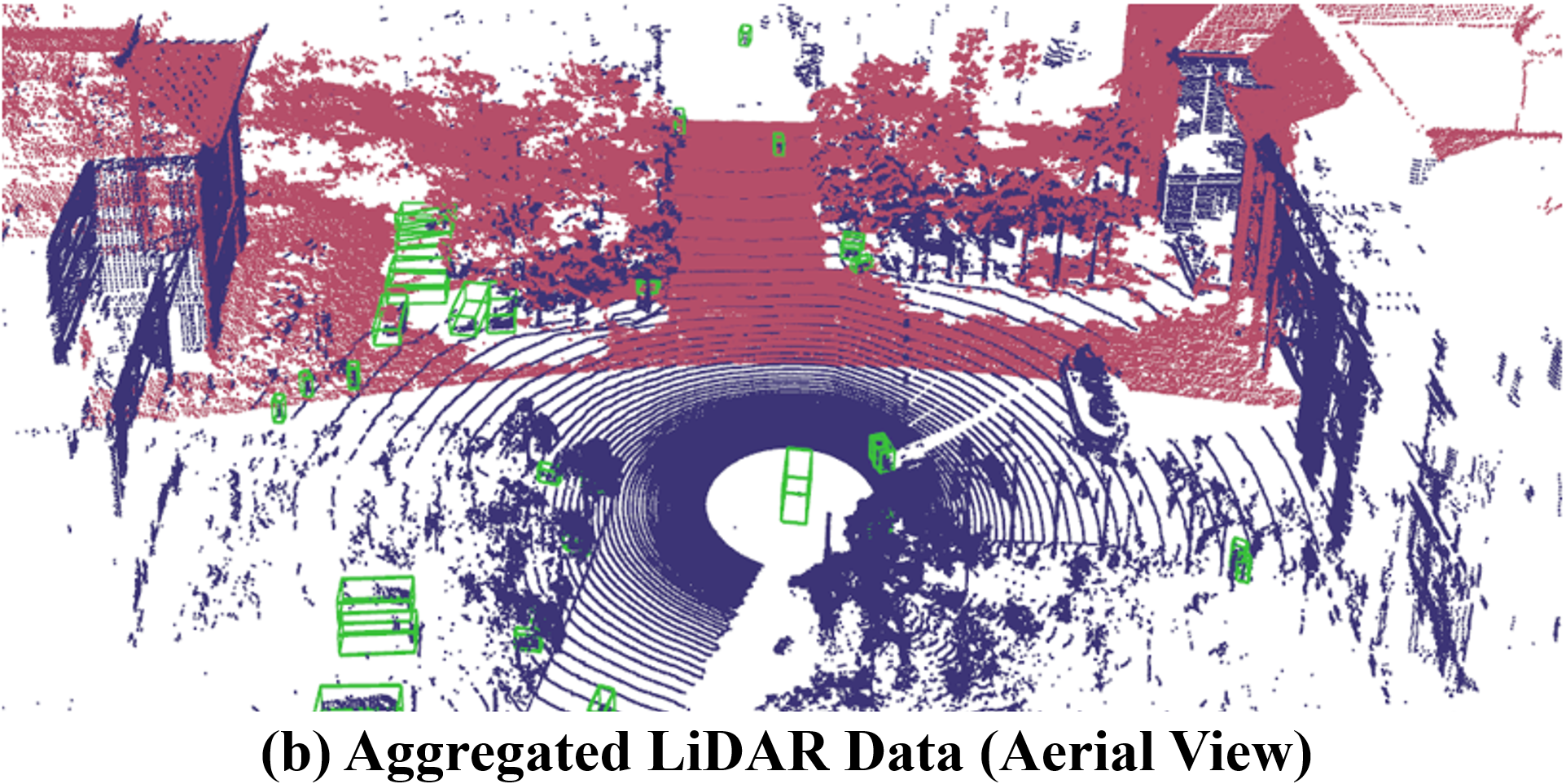}
        \end{minipage}
        \hfil
        \begin{minipage}{0.99\textwidth}
            \centering
            \includegraphics[width=\textwidth]{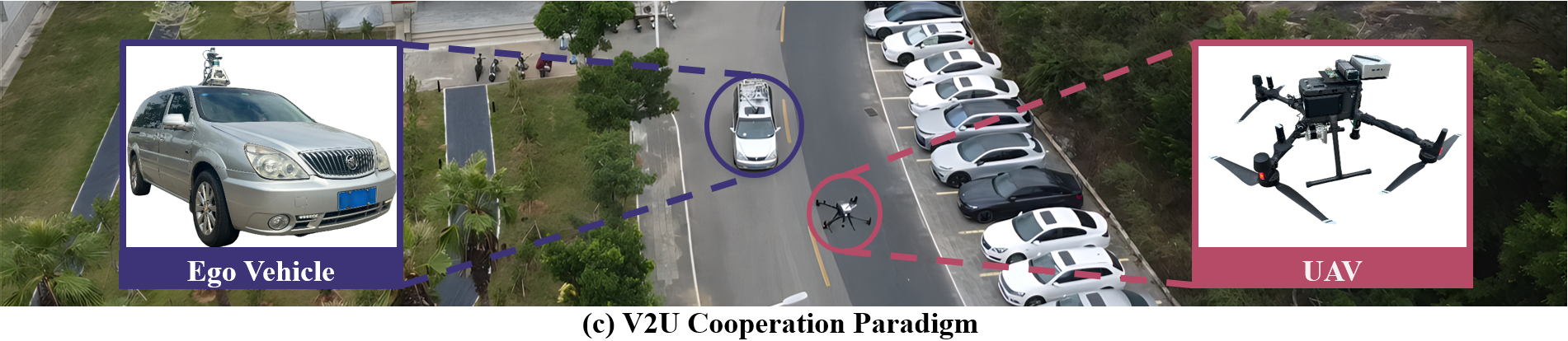}
        \end{minipage}

        \captionof{figure}{\textbf{An example data frame from V2U4Real.}
        (a) Ground-view aggregated LiDAR data.
        (b) Aerial-view aggregated LiDAR data.
        (c) V2U cooperation paradigm, where the ground vehicle serves as the ego.
        \textcolor{mypurple}{Purple points} are captured by the ground vehicle.
        \textcolor{myred}{Red points} are captured by the UAV.
        \textcolor{mygreen}{Green bounding boxes} are the ground truth.
        More qualitative examples are provided in the supplementary material.}
        \label{fig:data_sample}
    \end{center}
}]

\begingroup
\renewcommand\thefootnote{}
\footnotetext{* These authors contributed equally to this work.}
\footnotetext{$\dagger$ Corresponding Author: clwen@xmu.edu.cn}
\endgroup
\begin{abstract}Modern autonomous vehicle perception systems are often constrained by occlusions, blind spots, and limited sensing range. While existing cooperative perception paradigms, such as Vehicle-to-Vehicle (V2V) and Vehicle-to-Infrastructure (V2I), have demonstrated their effectiveness in mitigating these challenges, they remain limited to ground-level collaboration and cannot fully address large-scale occlusions or long-range perception in complex environments. To advance research in cross-view cooperative perception, we present \textbf{V2U4Real}, the first large-scale real-world multi-modal dataset for Vehicle-to-UAV (V2U) cooperative object perception. V2U4Real is collected by a ground vehicle and a UAV equipped with multi-view LiDARs and RGB cameras. The dataset covers urban streets, university campuses, and rural roads under diverse traffic scenarios, comprising over 56K LiDAR frames, 56K multi-view camera images, and 700K annotated 3D bounding boxes across four classes. To support a wide range of research tasks, we establish benchmarks for single-agent 3D object detection, cooperative 3D object detection, and object tracking. Comprehensive evaluations of several state-of-the-art models demonstrate the effectiveness of V2U cooperation in enhancing perception robustness and long-range awareness. The V2U4Real dataset and codebase is available at \href{https://github.com/VjiaLi/V2U4Real}{https://github.com/VjiaLi/V2U4Real}.\end{abstract}
  
\section{Introduction}

\label{sec:intro}

Cooperative perception is essential for autonomous systems, where multiple connected and automated vehicles (CAVs) share sensory information under the vehicle-to-vehicle (V2V) or vehicle-to-infrastructure (V2I) paradigms. This collaboration mitigates the limitations of single-vehicle perception, including occlusions, limited range, and sensor failures. Existing perception paradigms, such as the V2V paradigm (OPV2V~\cite{xu2022opv2v} and V2V4Real~\cite{xu2023v2v4real}) and the V2I paradigm (DAIR-V2X~\cite{yu2022dair}, V2X-Seq~\cite{yu2023v2x}, V2X-Sim~\cite{li2022v2x}, V2XSet~\cite{xu2022v2x}, and V2X-Real~\cite{xiang2024v2x} ), have significantly advanced this field. However, the V2V paradigm is constrained by the limited ground-level viewpoint coverage, while the V2I paradigm relies on costly fixed infrastructure.

\newcolumntype{C}{>{\centering\arraybackslash}X} 

\begin{table*}[t]
 \footnotesize 
 \centering
 \addtolength{\tabcolsep}{-3.8pt}
 \caption{\textbf{Comparisons of our proposed V2U4Real with existing cooperative perception datasets.} R/S: Real or Sim, V: Vehicle, U: UAV, I: Infrastructure, D: Detection, S: Segmentation, T: Tracking, \enquote*{\cmark}: Supported, \enquote*{\xmark}: Not Supported, \enquote*{-}: Not Applicable/Specified.}
 \begin{tabularx}{1.0\textwidth}{C|c|c|c|CC|CC|CCC|C}
  \toprule
\multirow{3}{*}{R/S} & \multirow{3}{*}{Dataset} & \multirow{3}{*}{Year} & \multirow{3}{*}{Paradigm}  & \multicolumn{2}{c|}{Vehicle Sensor} & \multicolumn{2}{c|}{UAV Sensor} &  \multicolumn{3}{c|}{Data Details} & \multirow{3}{*}{Annotations} \\
   \cmidrule(r){5-11}
 &  & &  & Camera & LiDAR & Camera & LiDAR & PointCloud & Image & 3D Boxes & \\  
\midrule
&OPV2V\cite{xu2022opv2v}&2022	&V2V  &\cmark &\cmark &- &-  &11K &44K &230K &D, T  \\
&V2X-Sim~\cite{li2022v2x}&2022	&V2V \& I  &\cmark &\cmark &- &- &10K &60K &26.6K &D, S, T \\
&V2XSet~\cite{xu2022v2x}&	2022	&V2V \& I &\cmark &\cmark &- &-  &11.4K &44K  &230K &D, T  \\
Sim&CoP-UAVs\cite{hu2022where2comm}&	2022 &U2U  &- &- &\cmark &\xmark &- &131.9K &1.94M &D, S \\
&UAV3D\cite{sunderraman2024uav3d}&	2024	&U2U  &- &- &\cmark &\xmark  &- &500K &3.3M &D, S, T  \\
&U2UData\cite{feng2024u2udata} &	2024	&U2U  &- &- &\cmark &\xmark  &315K &945K &2.41M &D, S, T  \\
&AirV2X\cite{gao2025airv2x}&	2025	&V2U\& I  &\cmark &\cmark &\cmark &\cmark  &121.1K &- &1.96M &D, S, T  \\
&Griffin\cite{wang2025griffin}&	2025	&V2U  &\cmark &\cmark &\cmark &\xmark  &37K &340K&- &D, T  \\
\midrule
&Dair-V2X\cite{yu2022dair}&2022 &V2V \& I  &\cmark &\cmark &- &- &39K &39K &464K &D, T \\
&V2V4Real\cite{xu2023v2v4real}&2023	&V2V  &\cmark &\cmark &- &- &20K &40K &240K &D, T\\
&V2X-Seq~\cite{yu2023v2x}&2023	&V2I  &\cmark &\cmark &- &-  &15K &15K &10.45K &D, T\\
&V2X-Real~\cite{xiang2024v2x}&2024	&V2V \& I  &\cmark &\cmark &- &-  &33K &171K &1.2M &D, T \\
Real&RCooper\cite{hao2024rcooper}&2024	&I2I  &- &- &- &-  &30K &50K &- &D, T \\
&TUMTraf-V2X\cite{zimmer2024tumtraf}&2024	&V2I  &\cmark &\cmark &- &- &2K &5K &29.38K &D, T\\
&CoPeD\cite{zhou2024coped}&2024 &V2U & \cmark &\cmark &\cmark &\xmark  &38K &203.4K &- &S  \\
&\textbf{V2U4Real(Ours)}&\textbf{2026} &\textbf{V2U}  &\cmark &\cmark &\cmark &\cmark &\textbf{56.2K} &\textbf{56.2K} &\textbf{702.8K} &\textbf{D, T} \\
\bottomrule
\end{tabularx}
\label{tab:dataset_comparison}
\end{table*}

Recently, the rapid adoption of unmanned aerial vehicles (UAVs) has enabled a flexible new perception paradigm, named Vehicle-to-UAV (V2U). Benefiting from its elevated global view and mobility, V2U complements the blind spots of ground vehicles, enabling wider coverage. As shown in Fig.~\ref{fig:data_sample}(a) and (b), in complex intersections, the ego vehicle (\textcolor{mypurple}{purple points}) suffers from severe occlusions, and the point clouds collected by a nearby UAV (\textcolor{myred}{red points}) significantly extend its perception range and reconstruct occluded objects. This capability is unattainable in previous paradigms.

\begin{figure}[!t]
\centering
\subfloat{%
  \includegraphics[width=.99\columnwidth]{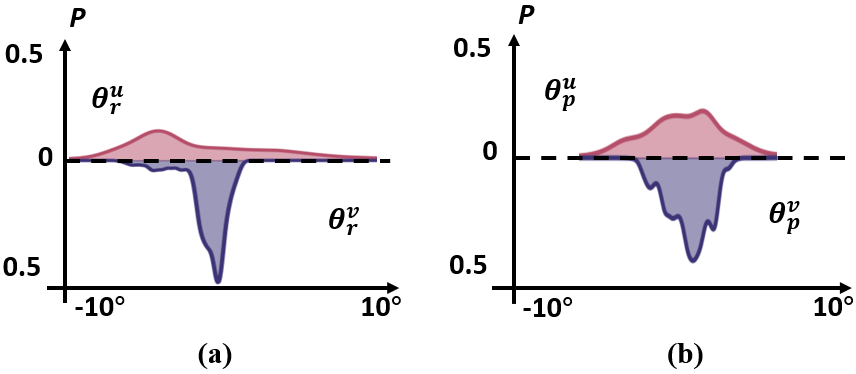}%
}
\caption{\textbf{Agent motion discrepancies between ground vehicle and UAV.}
(a) Probability distribution of the roll angle ($\theta_{r}$). 
(b) Probability distribution of the pitch angle ($\theta_{p}$).
\textcolor{myred}{Red color} denotes the UAV. \textcolor{mypurple}{Purple color} denotes the ego vehicle. $P$ denotes probability density.}

\label{fig:motivation}
\end{figure}

Despite its great potential, the V2U paradigm faces significant challenges arising from the inherent motion discrepancies between aerial and ground agents (Fig.~\ref{fig:motivation}(a) and (b)). UAVs experience full six-degrees-of-freedom (6-DoF) motion $(x, y, z, roll, pitch, yaw)$, resulting in a geometric mismatch between aerial and ground LiDAR data, which complicates coordinate alignment and cross-view fusion. Another major bottleneck lies in the scarcity of real-world large-scale V2U datasets. Most existing UAVs-related datasets, including CoP-UAVs~\cite{hu2022where2comm}, UAV3D~\cite{sunderraman2024uav3d}, Griffin~\cite{wang2025griffin} and AirV2X~\cite{gao2025airv2x}, are primarily generated through simulations using CARLA~\cite{carla_2017}, AirSim~\cite{shah2017airsim} and SUMO~\cite{sumo_2012}. These datasets lack realistic pose variations, occlusions, and dynamic interactions, which limit their applicability to real-world multi-view cooperative perception research. CoPeD~\cite{zhou2024coped} is a real-world ground-air dataset for cooperative perception, providing synchronized LiDAR and camera data for scenes such as forests and indoor settings; however, it only provides semantic annotations.

To further advance research on V2U cooperative perception, we present V2U4Real, the first large-scale real-world multi-modal dataset for cooperative object perception. V2U4Real covers urban streets, university campuses, and rural roads under diverse traffic scenarios. V2U4Real also integrates heterogeneous sensors such as mechanical spinning LiDAR, solid-state LiDAR, and RGB cameras, comprising over 56K LiDAR frames and 56K multi-view camera images. Additionally, V2U4Real provides 700K annotated 3D bounding boxes across four classes through multi-round manual data labeling, sensor calibration, and multi-source point cloud registration, enabling the validation of cooperative perception tasks. Compared to CoPeD~\cite{zhou2024coped}, V2U4Real covers more high-dynamic scenarios with diverse moving objects and heterogeneous sensors, thereby better representing complex traffic scenarios. Based on V2U4Real, we provide several benchmarks for both single-agent and multi-agent perception, including 3D object detection and object tracking, along with seven state-of-the-art cooperative perception algorithms. 

Our contributions can be summarized as follows:
\begin{itemize}
 \setlength{\parskip}{0pt}
  \setlength{\itemsep}{0pt plus 0pt}
    \item We build \textbf{V2U4Real}, the first large-scale real-world multi-modal dataset dedicated to V2U cooperative object perception, filling a critical gap in existing research on multi-view collaboration.
    \item We provide over $56K$ LiDAR frames, $56K$ camera images, and $700K$ manually annotated 3D bounding boxes across $4$ classes, all the frames captured by time-synchronized sensors in diverse real-world scenarios.
    \item We establish comprehensive benchmarks for both single-agent 3D object detection from distinct viewpoints and multi-agent cooperative detection and tracking. Additionally, we evaluate several state-of-the-art models to show the clear performance gains enabled by V2U cooperation.

\end{itemize}

\section{Related Works}

\subsection{Cooperative Perception Datasets}
Public datasets have significantly advanced research in cooperative perception recently. As summarized in Table~\ref{tab:dataset_comparison}, OPV2V~\cite{xu2022opv2v} leverages CARLA~\cite{carla_2017} and OpenCDA~\cite{xu2021opencda} co-simulation to construct the first 3D cooperative detection dataset. V2XSet~\cite{xu2022v2x} and V2X-Sim~\cite{li2022v2x} further investigate V2X perception using simulated environments. DAIR-V2X~\cite{yu2022dair} introduces the first real-world cooperative detection dataset but focuses exclusively on V2I collaboration, remaining limited to specific scenarios such as intersections. V2V4Real~\cite{xu2023v2v4real} provides a real-world V2V dataset, yet its perception coverage is inherently constrained by the ground-level perspective. V2X-Seq~\cite{yu2023v2x} extends DAIR-V2X~\cite{yu2022dair} by incorporating trajectory ID to support trajectory prediction, whereas V2X-Real~\cite{xiang2024v2x}, RCooper~\cite{hao2024rcooper}, and TUMTraf-V2X~\cite{zimmer2024tumtraf} offer multi-modal real-world datasets but remain predominantly vehicle-centric. Despite these advancements, cooperative perception datasets incorporating UAV viewpoints remain underdeveloped. Simulation-based datasets such as CoP-UAVs~\cite{hu2022where2comm}, UAV3D~\cite{sunderraman2024uav3d}, and U2UData~\cite{feng2024u2udata} primarily target UAV-to-UAV cooperation, while AirV2X~\cite{gao2025airv2x} and Griffin~\cite{wang2025griffin} further explore V2U cooperative perception but rely mainly on camera-only sensors, idealized communication and localization conditions, simplifying UAV dynamics under real-world physical constraints. CoPeD~\cite{zhou2024coped} pioneers a real-world ground–air cooperative system with automated semantic annotations, which mainly focuses on low-dynamic scenes, thereby limiting its applicability to complex traffic scenarios. The proposed V2U4Real is the first large-scale
real-world multi-modal dataset for V2U cooperative object perception.

\subsection{Single-agent Perception Algorithms}
Significant progress has been made in single-agent perception, where methods primarily focus on efficient feature extraction. Voxel-based approaches such as PointPillars~\cite{lang2019pointpillars} and SECOND~\cite{yan2018second} extract point cloud features via pillars or sparse convolutions for fast detection. CenterPoint~\cite{yin2021center} localizes object centers to accurately predict object dimensions and orientations, while PV-RCNN~\cite{shi2020pv} combines point-level and voxel-level features for high-precision detection in complex scenarios. Despite these advances, single-vehicle perception remains constrained by limited field-of-view (FOV) and sensing range, making distant or occluded objects challenging to detect.

\subsection{Cooperative Perception Algorithms}
Cooperative perception mitigates single-agent limitations by sharing information among multiple agents, which can be implemented via three primary fusion strategies: (1) \textbf{Early Fusion}~\cite{chen2019cooper}, where raw sensor data from all agents are aggregated for prediction; (2) \textbf{Late Fusion}~\cite{rawashdeh2018collaborative}, where individual detection outputs (\textit{e.g.}, 3D bounding boxes and confidence scores) are fused; (3) \textbf{Intermediate Fusion}, where intermediate features extracted from each agent are shared. Recent SOTA methods predominantly adopt intermediate fusion to balance accuracy and communication efficiency. AttFuse~\cite{xu2022opv2v} leverages self-attention mechanism to suppress noisy features, while Where2Comm~\cite{hu2022where2comm} selectively transmits salient task-relevant features. CoBEVT~\cite{xu2022cobevt} introduces a local–global sparse attention mechanism across agents, and V2X-ViT~\cite{xu2022v2x} employs a unified vision transformer for robust perception under GPS noise and latency. CoAlign~\cite{lu2023robust} further mitigates spatial misalignment. ERMVP~\cite{Zhang_2024_CVPR} improves robustness to localization noise and bandwidth limits via hierarchical sampling and consensus calibration, while DSRC~\cite{zhang2025dsrc} achieves density-insensitive and semantics-aware robustness through a distillation–reconstruction framework.

\section{V2U4Real Dataset}

In this section, we first describe the data acquisition process of V2U4Real in Sec.~\ref{sec:data_acquisition}, followed by the data annotation procedure in Sec.~\ref{sec:data_annotation}. Finally, we analyze the data distribution and present a detailed dataset analysis in Sec.~\ref{sec:data_analysis}.

\subsection{Data Acquisition}

\label{sec:data_acquisition}
\begin{figure}[!htb]
\hfil
\subfloat[Vehicle]{%
  \includegraphics[width=0.49\columnwidth]{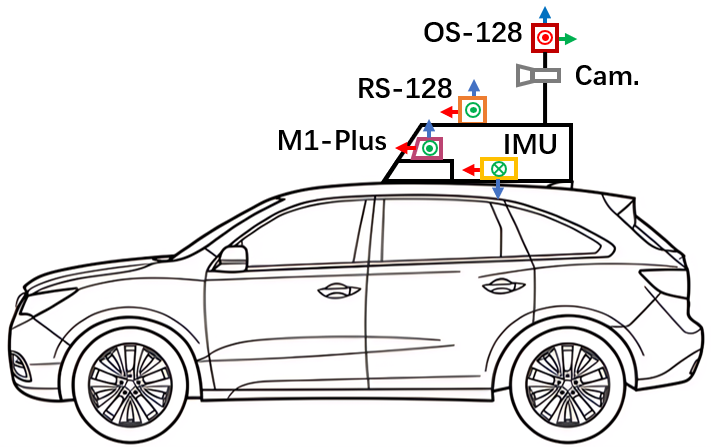}%
}
\hfil
\subfloat[UAV]{%
  \includegraphics[width=0.49\columnwidth]{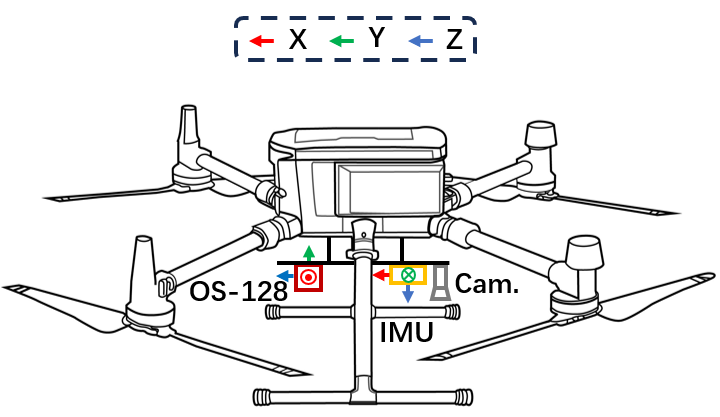}%
}
\caption{\textbf{Sensor overview of V2U4Real and coordinate of sensors.} The x,y, and z coordinates are red, green, and blue.}
\label{fig:platform}
\end{figure}

\noindent \textbf{Sensor Setup.}
V2U4Real is collected using two experimental connected perception agents, comprising a ground vehicle (Fig.~\ref{fig:platform}(a)) and a DJI M300 RTK UAV(Fig.~\ref{fig:platform}(b)). Both agents are equipped with LiDAR and RGB cameras. Specifically, the ego vehicle carries two mechanical spinning LiDARs (OS-128 and RS-128), one solid-state LiDAR (M1-Plus), and three RGB cameras (left, center, and right). The UAV carries a single mechanical spinning LiDAR (OS-128) and one downward-facing camera. Moreover, each agent integrates a high-precision GPS/IMU system for accurate localization, enabling the initial spatial alignment of LiDAR point clouds collected by the vehicle and UAV. The overall sensor configuration is illustrated in Fig.~\ref{fig:platform}, with detailed sensor specifications provided in Tab.~\ref{tab:sensor_specifications}.

\begin{figure}[!htb]
    \centering
    \includegraphics[width=1\columnwidth]{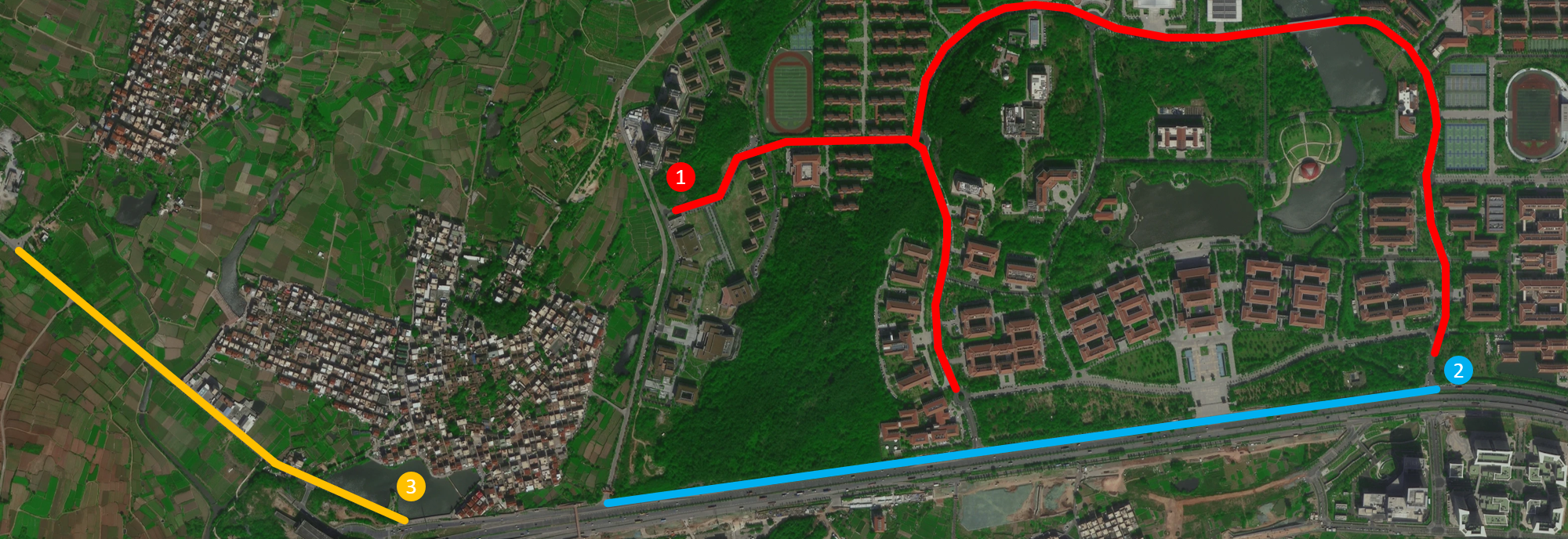}
    \caption{\textbf{Driving routes of two data collection agents.} Urban, rural, and campus roads are represented by blue, yellow, and red lines, respectively.}
    \label{fig:driving_route}
\end{figure}

\noindent\textbf{Driving Route.} During data collection, the two agents move simultaneously, maintaining a horizontal distance within 100 meters, while the UAV flies at 50 meters above ground to ensure effective collaborative perception. To enhance multi-view cooperation, we vary the relative poses ($\theta_r$, $\theta_p$, $\theta_y$) of the agents during data collection (see Fig.~\ref{fig:motivation}(a) and (b) for details). Our dataset covers three representative driving scenarios: urban roads, rural roads, and campus roads. The corresponding routes are illustrated in ~\cref{fig:driving_route}.

\noindent\textbf{Data Collection.} We conducted data collection with all sensors recording at 10 Hz under a unified temporal reference. After collection, we manually selected the 44 most representative scenarios, excluding any data loss or anomalies caused by external factors. Each scenario lasts between 30 and 80 seconds, generating approximately 56,200 frames of LiDAR point clouds and RGB images.

\begin{table*}[h!t]
\small
\centering
 \caption{\textbf{Key Sensor Specifications in V2U4Real Dataset.}}
\begin{tabular}{l|l|l|p{7.8cm}}
\toprule
Platform&Sensor&Sensor Model & Details  \\
\toprule
\multirow{8}{*}{Vehicle}
&LiDAR& Ouster-1-128($\times$1) & 128 channels, 10Hz capture frequency, $360^\circ$ horizontal FOV, $-22.5^\circ \text{to} +22.5^\circ$ vertical FOV, $\leq$200m range;\\
&LiDAR& RoboSense Ruby-Plus-128 ($\times$1) & 128 channels, 10Hz capture frequency, $360^\circ$ horizontal FOV, $-25^\circ \text{to} +15^\circ$ vertical FOV, $\leq$250m range;\\
&LiDAR& RoboSense M1 Plus($\times$1) & 10Hz capture frequency, $120^\circ$ horizontal FOV, $-12.5^\circ \text{to} +12.5^\circ$ vertical FOV, $\leq$200m range;\\
&Camera & HikRobot MV-CA023-10GC ($\times$3) & RGB, 10Hz capture frequency, 1920×1080 resolution; \\ 
&GPS/IMU & Quectel LC29H ($\times$1) / Sigma-D4G ($\times$1) & 1000Hz update rate, RTK supported, Double-Precision; \\ 
\midrule
\multirow{4}{*}{UAV}
&LiDAR& Ouster-1-128 ($\times$1) & 128 channels, 10Hz capture frequency, $360^\circ$ horizontal FOV, $-22.5^\circ \text{to} +22.5^\circ$ vertical FOV, $\leq$200m range;\\
&Camera & HikRobot MV-CA023-10GC  ($\times$1) & RGB, 10Hz capture frequency, 1920×1080 resolution; \\ 
&GPS/IMU& Quectel LC29H ($\times$1) / DETA100-D4G ($\times$1) & 1000Hz update rate, RTK supported, Double-Precision;\\ 
\bottomrule
\end{tabular}
\label{tab:sensor_specifications}  
\end{table*}

\begin{figure*}[!t]
\centering
\subfloat[Calibration results between Camera and LiDAR]{%
  \includegraphics[width=1\textwidth]{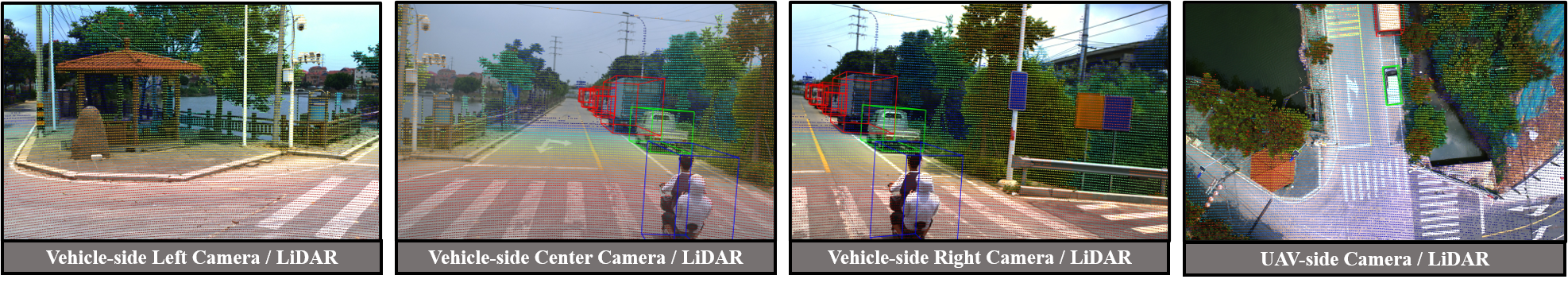}%
}

\subfloat[Point cloud registration results between ego vehicle and UAV]{%
  \includegraphics[width=1\textwidth]{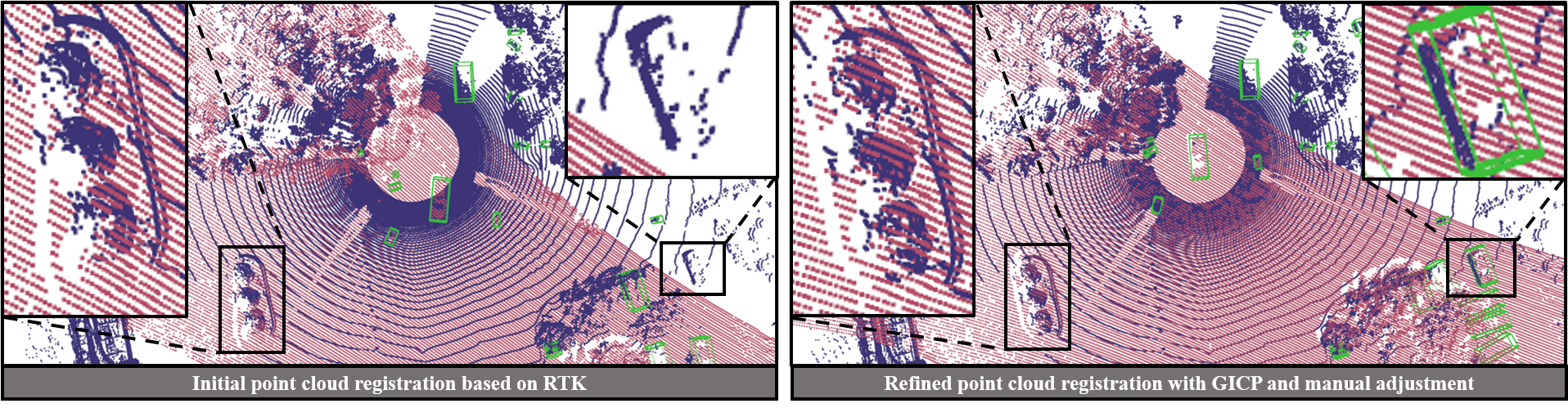}%
}

\caption{\textbf{Visualization of sensor calibration and point cloud registration results.} \textcolor{mypurple}{Purple points} are captured by the ground vehicle. \textcolor{myred}{Red points} are captured by the UAV. \textcolor{mygreen}{Green bounding box} is the ground truth.}
\label{fig:Calibration_and_Registration}
\vspace{-2mm}
\end{figure*}

\noindent \textbf{Sensor Calibration and Registration.}
Spatial synchronization of the camera and LiDAR is performed through the sensor calibration process \cite{zhang2002flexible, geiger2012automatic, koide2021voxelized}. The intrinsic parameters of the camera are calibrated using a checkerboard pattern, while the LiDAR is calibrated relative to the camera by utilizing 5 pairs of points extracted from the point cloud and the corresponding camera image for each frame. The extrinsic parameters are obtained by minimizing the reprojection errors between the 2D-3D point correspondences. The results of this calibration are visually represented in Fig.~\ref{fig:Calibration_and_Registration}(a). LiDAR alignment between the vehicle and UAV platforms is achieved through point cloud registration, initially computed using RTK localization and subsequently refined via GICP~\cite{koide2021voxelized} and manual adjustments. The visualization of the point cloud registration is shown in Fig.~\ref{fig:Calibration_and_Registration}(b).

\subsection{Data Annotation}
\label{sec:data_annotation}

\noindent\textbf{Coordinate System.} Our dataset adopts three different coordinate systems: (1) the LiDAR local coordinate system. Although the LiDAR sensors on each agent are mounted in different directions, their local coordinate systems are unified through a predefined transformation matrix, where the X, Y, and Z axes correspond to the forward, left, and upward directions, respectively. (2) the camera coordinate system, in which the Z axis denotes depth. (3) the world coordinate system, following the North-East-Down (NED) convention. 3D bounding boxes are annotated separately in each LiDAR’s local coordinate system, allowing the data from each LiDAR also to be used for single-agent perception tasks. We utilize the positional information provided by GPS to initialize their relative poses for each frame.

\noindent\textbf{3D Bounding Box Annotation.} We employ SusTechPoint~\cite{li2020sustech}, a powerful opensource labeling tool, to annotate 3D bounding boxes for the collected LiDAR data. Two groups of professional annotators are involved. One group is responsible for the initial labeling, while the other group further refines the annotations. There are four object classes, including Car, Cyclist, Pedestrian, and Truck. For each object, we annotate its 3D bounding box, which contains $(x, y, z)$ for the central position, ($l, w, h$) for the bounding box extent, and $(roll, pitch, yaw)$ for Euler angles. To facilitate downstream perception tasks, such as tracking and behavior prediction, we assign a consistent ID to the same object across different timestamps and record the moving velocity of each object. 

All 3D bounding boxes are annotated independently in each LiDAR’s local coordinate system; however, two issues may occur: (1) the same object may be assigned a different ID across sensors. (2) an object may be annotated in one sensor but be missing in the other. To address these issues, we transform all LiDAR point clouds and 3D bounding boxes into the ego LiDAR local coordinate system. Then, we perform IoU matching between 3D bounding boxes from different LiDAR sensors. If the IoU exceeds a predefined threshold, which is category-dependent to reflect different object sizes, inconsistent IDs are reassigned. Additionally, unannotated instances with more than five points are re-labeled to ensure consistency across LiDAR sensors.

\subsection{Data Analysis}
\label{sec:data_analysis}

\noindent\textbf{Agent Motion Discrepancies.} As shown in Fig.~\ref{fig:motivation}, vehicle-mounted LiDAR sensors exhibit highly stable motion, with roll ($\theta_r$) and pitch ($\theta_p$) variations mostly confined within $\left|\theta_r\right|, \left|\theta_p\right| \le 2^\circ$, where the variations refer to the inter-frame change $\Delta_{\theta}$ between consecutive frames. In contrast, UAV experience substantially larger agent motion fluctuations, with $\left|\theta_r\right|, \left|\theta_p\right| \le 10^\circ$, reflecting their higher maneuverability. This stark contrast underscores the stability of ground vehicles versus the agility of UAV, posing additional challenges for cross-agent point cloud alignment and fusion.

\begin{figure}[!t]
\centering
\subfloat{%
  \includegraphics[width=\columnwidth]{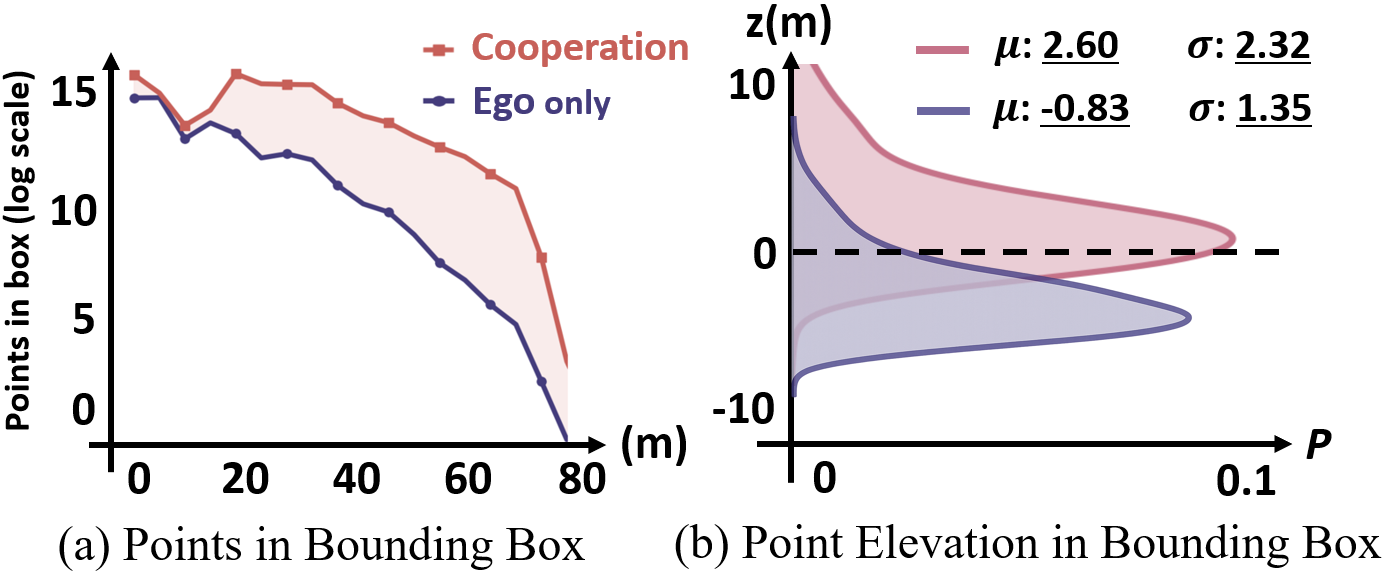}%
}
\caption{\textbf{Data analysis of V2U4Real.} $P$ denotes probability, $\mu$ denotes mean, and $\sigma$ denotes standard deviation. }
\label{fig:Data_analysis}
\vspace{-4mm}
\end{figure}

\noindent\textbf{Points in Bounding Box.} Fig.~\ref{fig:Data_analysis}(a) shows the number of LiDAR points (log-scaled) within the bounding boxes with respect to the radial distance from the ego vehicle. When only the ego vehicle performs perception, the number of LiDAR points decreases dramatically as the distance increases. In contrast, with cooperative perception, the point density remains consistently higher, especially in long-range regions beyond 50 m. This demonstrates that cooperation effectively mitigates occlusion and sparsity, significantly enhancing perceptual coverage.

\noindent\textbf{Point Elevation in Bounding Box. }
Beyond the roll and pitch jitter induced by agent motion, the overall elevation distribution $P_z$ of points within bounding boxes also varies notably across agents due to differences in their operating heights and sensing perspectives. As shown in Fig.~\ref{fig:Data_analysis}(b), most points from the ego vehicle are concentrated near its own height ($\mu = -0.83$, $\sigma = 1.35$), whereas points from the UAV, owing to its bird’s-eye view, are distributed at considerably higher elevations ($\mu = 2.60$, $\sigma = 2.32$).

\section{Tasks}
V2U4Real supports multiple downstream perception tasks, including single-agent and cooperative detection, tracking, prediction, and localization. In this paper, we focus on single-agent and cooperative detection and tracking tasks.

\subsection{Single-agent 3D Object Detection}
\label{sec:single_detection}
\paragraph{Task Definition.}
Single-agent 3D object detection involves two tasks: vehicle-side detection and UAV-side detection. Each agent performs training and inference solely based on the data collected by its onboard sensors.

\noindent\textbf{Groundtruth.} As described in Sec.~\ref{sec:data_annotation}, 3D bounding boxes are annotated in the local coordinate system of each LiDAR sensor. For convenience in training and evaluation, point clouds and 3D bounding boxes from the UAV-mounted LiDAR are transformed into the local coordinate frame of the ego LiDAR via a predefined transformation matrix. Then, each agent performs training and evaluation based on its respective annotations within this unified coordinate frame.

\noindent\textbf{Evaluation.} The evaluation range along the $x$ and $y$ axes is $[-100, 100]$ m and $[-80, 80]$ m, respectively. As shown in Fig.~\ref{fig:Data_analysis}(a), since most traffic participants are vehicles and cyclists, this task primarily focuses on Vehicle and Cyclist detection. Following~\cite{xu2022v2x, xu2023v2v4real}, we group Car and Truck into Vehicle category. Consistent with KITTI~\cite{geiger2012we}, we adopt Average Precision (AP) at IoU thresholds of 0.5 and 0.7 for Vehicle, and 0.25 and 0.5 for Cyclist.

\noindent\textbf{Benchmarking Methods.} We benchmarked several representative LiDAR-based 3D object detection models, including PointPillars~\cite{lang2019pointpillars}, SECOND~\cite{yan2018second}, CenterPoint~\cite{yin2021center}, and PV-RCNN~\cite{shi2020pv}.

\newcolumntype{C}{>{\centering\arraybackslash}X} 

\begin{table*}[!t]
\footnotesize
\centering
\caption{\textbf{Cooperative 3D object detection benchmarks for vehicle category on V2U4Real.} Sync. means synchronous setup ignoring communication delays. Async. implies asynchronous setup with a $(0, 1000]$\,ms delay. The vehicle category includes car and truck.}
\setlength{\tabcolsep}{3pt}
\begin{tabularx}{\textwidth}{c|c|c|CCC|CCC|C}
\toprule
\multirow{3}{*}{Paradigm} & \multirow{3}{*}{Method} & \multirow{3}{*}{Year} & 
\multicolumn{3}{c|}{Sync. (3D AP@IoU = 0.5 / 0.7)} & 
\multicolumn{3}{c|}{Async. (3D AP@IoU = 0.5 / 0.7)} & 
\multicolumn{1}{c}{AM} \\
\cmidrule(r){4-9}
& & &Overall & 0–50m & 50–100m & Overall & 0–50m & 50–100m & (MB)\\
\midrule

\multirow{9}{*}{Vehicle to UAV}
& Early Fusion & - & 51.31/30.97 & 57.53/35.67 & 24.47/14.99 & 30.94/13.99 & 34.17/15.85 & 20.33/8.91 & 3.18 \\
& Late Fusion & - & 43.61/27.74 & 49.28/31.73 & 16.75/8.56 & 28.08/16.18 & 29.63/17.31 & 13.87/6.56 & 0.009\\
& Where2comm~\cite{hu2022where2comm} & 2022 & 53.85/29.71 & 60.47/33.14 & 26.73/16.59 & 48.99/28.36 & 55.30/31.83 & 25.51/14.46 & 0.65 \\
& V2X-ViT~\cite{xu2022v2x}& 2022 & 45.51/25.99 & 50.98/29.65 & 19.67/12.24 & 45.74/26.00 & 51.29/29.68 & 19.65/12.13 &  0.65 \\
& AttFuse~\cite{xu2022opv2v} & 2022 & 50.20/27.33 & 55.31/30.41 & 29.33/18.39 & 40.51/21.92 & 44.71/23.41 & 28.51/16.98 & 0.65  \\
& CoBEVT~\cite{xu2022cobevt} & 2022 & 41.99/22.38 & 48.46/26.40 & 21.58/13.49 & 31.47/16.05 & 36.16/18.67 & 20.60/12.83 & 0.65 \\
& CoAlign~\cite{lu2023robust}& 2023  & \textbf{56.67/36.61} & \textbf{63.21/41.37} & 30.20/19.25 & \textbf{50.81/33.33} & \textbf{56.58/37.71} & 29.65/18.14 & 0.65 \\
& ERMVP~\cite{Zhang_2024_CVPR} & 2024 & 45.56/22.64 & 51.89/26.69 & 23.09/12.91 & 28.06/14.22 & 31.58/16.22 & 21.15/11.93 & 0.65 \\
& DSRC~\cite{zhang2025dsrc} & 2025 & 54.64/31.77 & 59.61/34.41 & \textbf{33.26/20.63} & 47.63/26.05 & 51.50/28.46 & \textbf{31.22/19.02} & 0.65 \\
\midrule
Vehicle only & No Fusion &- & 27.53/12.75 & 40.12/20.43 & 15.54/6.23 & 27.53/12.75 & 40.12/20.43 & 15.54/6.23 & 0 \\
\midrule
UAV only & No Fusion &- & 32.44/14.31 & 36.71/16.42 & 19.74/10.11 & 32.44/14.31 & 36.71/16.42 & 19.74/10.11 & 0 \\
\bottomrule
\end{tabularx}
\label{tab:exps_coperception}
\end{table*}

\subsection{Cooperative 3D Object Detection}
\noindent\textbf{Task Definition.} 
The V2U4Real 3D cooperative object detection task involves fusing the point clouds captured by the UAV with those from the ego vehicle to perform 3D object detection. Compared to single-agent detection, cooperative detection presents several domain-specific challenges:
\begin{itemize}
    \item \textbf{Spatial Asynchrony:} Agents observe the same scene from distinct viewpoints, causing varying FOV overlap and occlusion patterns. Moreover, the 6-DoF motion of UAV, coupled with GPS errors, leads to drift when transforming UAV point clouds into the ego frame.

    \item \textbf{Temporal Asynchrony:} Differences in sensor rates, clock drift, and communication latency introduce temporal misalignment among agents. Limited bandwidth restricts the exchange of intermediate features, thereby reducing the consistency and efficiency of multi-agent fusion, especially in dynamic, rapidly changing scenes.

\end{itemize}
The primary mission of this task is to develop effective cooperative detection methods that can address the challenges above.

\noindent\textbf{Groundtruth.} During both training and evaluation, the ground vehicle is always selected as the ego vehicle, and the UAV transforms its LiDAR data and annotated 3D bounding boxes into the ego’s LiDAR local coordinate system. Unlike single-agent detection, this task utilizes point clouds and ground truth from all agents throughout the cooperative perception pipeline, which is formulated as follows:
\begin{equation}
GT = GT_v \cup GT_u,
\label{eq:gt_merge}
\end{equation}
where $GT_v$ and $GT_u$ denote the ground truth for the vehicle-side sensor and the UAV-side sensor perception, respectively.
Due to spatial and temporal asynchrony, the 3D bounding boxes from each agent corresponding to the same object may exhibit offsets. In such a case, the annotations from the ego vehicle would be retained as the ground truth.

\noindent\textbf{Evaluation.} The evaluation range along the $x$ and $y$ axes is $[-15, 100]$ m and $[-80, 80]$ m, respectively, as we focus on forward-facing cooperative perception of the ego vehicle, where interactions are more frequent and safety-critical. To assess transmission costs, Average MegaByte (AM) is employed, representing the size of transmitted data required by each algorithm. Following~\cite{yu2022dair, xu2023v2v4real}, we evaluate all models under two settings: 1) \textbf{Sync setting}, ignoring communication delays; 2) \textbf{Async setting}, where a realistic communication delay of $(0, 1000]$\,ms is simulated by retrieving the LiDAR data from the previous timestamp. 

\noindent\textbf{Benchmarking Methods.} We present comprehensive benchmarks for four representative fusion strategies in cooperative 3D object detection:

\begin{itemize}
    \item \textbf{No Fusion:} The models are trained independently on each agent using their own LiDAR data.
However, during the evaluation, a shared ground truth space (as defined in Eq.~\ref{eq:gt_merge}) is adopted.
    \item \textbf{Late Fusion:} Each agent independently performs 3D object detection using its own LiDAR data and transmits the predictions to the ego vehicle. The ego vehicle then applies NMS~\cite{neubeck2006efficient} to generate the final detection results.
    \item \textbf{Early Fusion:} All point clouds are transformed into the ego vehicle's LiDAR local coordinate system, and detection is subsequently performed on the fused point cloud.
    \item \textbf{Intermediate Fusion:} Each agent extracts intermediate point cloud features via a backbone network, which are then compressed and transmitted to the ego vehicle for feature fusion and joint detection. We evaluate several leading methods, including AttFuse~\cite{xu2022opv2v}, Where2comm~\cite{hu2022where2comm}, V2X-ViT~\cite{xu2022v2x}, CoBEVT~\cite{xu2022cobevt}, CoAlign~\cite{lu2023robust}, ERMVP~\cite{zhang2024ermvp} and DSRC~\cite{zhang2025dsrc} to establish a solid benchmark for this task.
\end{itemize}

\subsection{Cooperative 3D Object Tracking}
\noindent\textbf{Task Definition.} This task aims to investigate how cooperative object detection can enhance the accuracy of object tracking. Two principal paradigms exist in multiple object tracking: 1) joint detection and tracking; 2) tracking by detection. Our study focuses on the latter.

\noindent\textbf{Evaluation.} 
Following~\cite{caesar2020nuscenes, weng20203d, xu2023v2v4real}, we evaluate 3D object tracking performance using seven commonly adopted metrics: 1) Multi Object Tracking Precision (MOTA), 2) Multi Object Tracking Accuracy (MOTP), 3) Mostly Tracked (MT), 4) Mostly Lost (ML), 5) Average MOTA (AMOTA), 6) Average Multi Object Tracking Precision (AMOTP), and 7) scaled AMOTA (sAMOTA). AMOTA and AMOTP extend the traditional MOTA and MOTP by averaging their values across all recall thresholds, thereby reflecting the distribution of confidence in predictions. The sAMOTA metric~\cite{caesar2020nuscenes} further normalizes this measure to ensure a linear evaluation across the entire recall range, facilitating fairer comparisons in complex tracking scenarios.

\noindent\textbf{Baselines Tracker.} We employ the AB3Dmot tracker~\cite{weng2020ab3dmot} as our baseline. It extends the classical SORT~\cite{Bewley_2016} framework to 3D space by integrating a 3D Kalman Filter~\cite{kalman1960new} for motion prediction and the Hungarian algorithm~\cite{kuhn1955hungarian} for data association, achieving efficient and robust tracking based on cooperative detection results.
\begin{table}[t]
\footnotesize
\centering
\addtolength{\tabcolsep}{-2pt}
\caption{\textbf{Single-agent 3D object detection benchmarks on V2U4Real.} The vehicle category includes car and truck.}
\begin{tabularx}{0.48\textwidth}{c|c|CC|CC}
\toprule
\multirow{2}{*}{Platform} & \multirow{2}{*}{Method}  & \multicolumn{2}{c|}{\text{Veh. (3D AP@IoU)}} & \multicolumn{2}{c}{\text{Cyc. (3D AP@IoU)}} \\ 
\cmidrule(lr){3-6}
& & 0.5 & 0.7 & 0.25 & 0.5 \\
\midrule
\multirow{4}{*}{Vehicle}
& PointPillars~\cite{lang2019pointpillars} & 67.06 & 30.22 & \textbf{60.45} & 50.12 \\
& SECOND~\cite{yan2018second} & 60.64 & 30.01 & 57.67 & 49.98 \\
& CenterPoint~\cite{yin2021center} & 65.79 & 26.96 & 59.89 & 50.48\\
& PV-RCNN~\cite{shi2020pv} & \textbf{68.18} & \textbf{36.23} & 59.51 & \textbf{51.31} \\
\midrule
\multirow{4}{*}{UAV}
& PointPillars~\cite{lang2019pointpillars} & 70.09 & 47.06 & 57.33 & 41.61 \\
& SECOND~\cite{yan2018second} & 66.67 & 48.96 & 57.35 & 45.65 \\
& CenterPoint~\cite{yin2021center} & 70.01 & 49.20 & 63.51 & 49.67 \\
& PV-RCNN~\cite{shi2020pv} & \textbf{72.26} & \textbf{55.15} & \textbf{64.28} & \textbf{50.86} \\
\bottomrule
\end{tabularx}
\label{tab:exps_detection_single}
\vspace{-1mm}
\end{table}

\newcolumntype{C}{>{\centering\arraybackslash}X}

\begin{table*}[t]
 \footnotesize 
 \centering
 \addtolength{\tabcolsep}{-3.8pt}
 \caption{\textbf{3D object tracking benchmarks for vehicle category on V2U4Real.}}
 \begin{tabularx}{1.0\textwidth}{c|c|c|CCCCCCC}
  \toprule
 Paradigm &Method & Year & AMOTA(↑) & AMOTP(↑) & sAMOTA(↑) & MOTA(↑) & MOTP(↑) & MT(↑) & ML(↓)\\  
  \midrule
\multirow{9}{*}{Vehicle to UAV}  
&Early Fusion & - &19.22 & 39.41 & 56.79 & 59.26 & 64.67 & 67.94 & 23.81 \\
&Late Fusion  & - &14.82 &34.39 &51.12 &50.64 &64.90 &48.41 & 37.30\\
&AttFuse\cite{xu2022opv2v} & 2022 & 20.98 & 41.31 & \textbf{61.55} & 60.40  & 64.89 & 57.94 & 23.81 \\
&Where2comm\cite{hu2022where2comm} & 2022 &20.07 & 41.49 & 58.52 & 62.07 & 63.04 & 65.08 & 15.87 \\
&V2X-ViT~\cite{xu2022v2x}& 2022 & 14.74 & 35.25 & 50.38 & 50.76 & 64.76 & 57.94 & 30.16 \\
&CoBEVT\cite{xu2022cobevt}& 2022 & 15.96 & 36.53 & 52.62 & 53.60  & 64.54 & 51.59 & 35.71  \\
&CoAlign~\cite{lu2023robust} & 2023 & \textbf{22.08} & \textbf{43.11} & 59.03 & \textbf{63.49} & \textbf{65.43} & 69.05 & 14.29 \\
&ERMVP~\cite{Zhang_2024_CVPR} & 2024 & 19.02 & 37.42 & 56.82 & 58.27 & 63.31 & 35.71 & 49.21 \\
&DSRC\cite{zhang2025dsrc} & 2025 & 18.67 & 39.72 & 53.20 & 59.74 & 65.03 & \textbf{73.02}  & \textbf{13.49}   \\
\midrule
Vehicle only&No Fusion  & - & 11.73 & 25.84 & 46.56 & 45.33 & 44.59 &  34.13 & 52.38  \\
\midrule
UAV only& No Fusion & - & 7.00 & 21.58 & 33.56 & 35.72 & 55.40 &  3.17 & 81.75  \\
\bottomrule
\end{tabularx}
\label{tab:exps_tracking}

\end{table*}

\section{Experiments}
\subsection{Implementation Details}
V2U4Real is divided into train/val/test sets with a ratio of 4:1:1 for all three tasks. All experimental results are evaluated on the val set. We choose OS-128 for the main experiments. The remaining results, including those on the test set and those obtained with other sensors, are provided in the supplementary material. For the cooperative detection task, all models adopt PointPillar~\cite{lang2019pointpillars} as the backbone to extract 2D features from point clouds. All models are trained for 40 epochs with a batch size of 4 per RTX 3090 GPU and an initial learning rate of 0.002. The learning rate is decayed using a cosine annealing schedule~\cite{loshchilov2019decoupledweightdecayregularization}, and early stopping is applied to select the optimal model. Standard point cloud augmentations are used in all experiments. We employ the AdamW~\cite{kingma2014adam} optimizer with a weight decay of $1\times10^{-2}$.

\subsection{Benchmark Results Analysis}

\noindent\textbf{Single-agent 3D Object Detection.}
Tab.~\ref{tab:exps_detection_single} shows that single-agent 3D object detection performed on the UAV generally outperforms its counterpart on the ego vehicle across all IoU thresholds, especially for the Vehicle category, achieving 5\%/20\% higher AP@IoU=0.5/0.7. This superiority arises from the UAV’s elevated viewpoint, which provides a less occluded view of surrounding scenes. In contrast, the ego vehicle suffers from frequent occlusions and limited vertical FOV, with most methods fluctuating around 30\% AP@IoU=0.7. However, for the Cyclist category, the UAV’s bird’s-eye perspective becomes a double-edged sword. While it offers a comprehensive view of the scene, it also results in sparse point distributions for small-scale objects, making precise detection more challenging.

\noindent\textbf{Cooperative vs. No Fusion.} 
Tab.~\ref{tab:exps_coperception} shows that cooperative methods substantially outperform the No Fusion baselines under both synchronous and asynchronous settings. The Vehicle-only baseline achieves 27.53\%/12.75\% AP@IoU=0.5/0.7, while the UAV-only baseline performs slightly with 32.44\%/14.31\% AP@IoU=0.5/0.7. Among all cooperative methods, CoAlign~\cite{lu2023robust} achieves the best overall performance, reaching 56.67\%/36.61\% AP@IoU=0.5/0.7 under the synchronous setting, and also maintains the highest performance in the asynchronous setting with 50.81\%/33.33\% AP@IoU=0.5/0.7. This demonstrates its robustness to spatial misalignment. 

\noindent\textbf{Advantages of Collaboration at Long Range.}
Tab.~\ref{tab:exps_coperception} shows that cross-agent feature exchange effectively extends the perception horizon, achieving up to 10\% AP@IoU=0.5 over the No Fusion baselines for targets at 50 to 100 meters. DSRC~\cite{zhang2025dsrc} achieves the best performance with 33.26\%/20.63\% and 31.22\%/19.02\% AP@IoU=0.5/0.7 under synchronous and asynchronous settings at long range due to semantic-guided reconstruction.

\noindent\textbf{Temporal Sync. vs. Async.}
Tab.~\ref{tab:exps_coperception} shows that introducing communication latency leads to performance drops for all cooperative methods, particularly under strict IoU thresholds. Despite this, CoAlign~\cite{lu2023robust} maintains the highest resilience, with 50.81\%/33.33\% AP@IoU=0.5/0.7. V2X-ViT~\cite{xu2022v2x} achieves slight performance improvements by leveraging temporal compensation mechanisms that mitigate feature misalignment caused by asynchronous updates. Early and Late Fusion baselines are more affected, indicating a limited ability to handle temporal misalignment.

\noindent\textbf{Cooperative 3D Object Tracking.}
Tab.~\ref{tab:exps_tracking} shows the benchmark results for cooperative 3D object tracking task. Similar to the cooperative 3D object detection task, cooperative 3D object tracking significantly outperforms the No Fusion baselines. CoAlign~\cite{lu2023robust} achieves the best performance across most metrics, including AMOTA (22.08\%, which is 10.35\% higher than the Vehicle-only baseline) and AMOTP (43.11\%, which is 17.27\% higher than the Vehicle-only baseline). DSRC~\cite{zhang2025dsrc} also improves notably, achieving 73.02\% MT and 13.49\% ML.

\section{Conclusion}
In this work, we present V2U4Real, the first large-scale real-world multi-modal dataset for Vehicle-to-UAV cooperative object perception. It fills an important domain gap in existing research by enabling more effective multi-view collaboration between UAV and the ego vehicle. V2U4Real contains 56K LiDAR frames, 56K multi-view camera images, and 700K manually annotated 3D bounding boxes. To support a wide range of research tasks, V2U4Real offers well-structured benchmarks for single-agent 3D object detection, cooperative 3D object detection, and cooperative 3D object tracking, providing a unified platform for studying Vehicle-to-UAV cooperative object perception.

\noindent \textbf{Acknowledgement.} This work was supported by the National Natural Science Foundation of China (No.42571514).

\vspace{0.5\baselineskip}
\newpage
\vspace{-\baselineskip}
{
    \small
    \bibliographystyle{ieeenat_fullname}
    \bibliography{main}

\begin{thebibliography}{44}
\providecommand{\natexlab}[1]{#1}
\providecommand{\url}[1]{\texttt{#1}}
\expandafter\ifx\csname urlstyle\endcsname\relax
  \providecommand{\doi}[1]{doi: #1}\else
  \providecommand{\doi}{doi: \begingroup \urlstyle{rm}\Url}\fi

\bibitem[Bewley et~al.(2016)Bewley, Ge, Ott, Ramos, and Upcroft]{Bewley_2016}
Alex Bewley, Zongyuan Ge, Lionel Ott, Fabio Ramos, and Ben Upcroft.
\newblock Simple online and realtime tracking.
\newblock In \emph{2016 IEEE International Conference on Image Processing (ICIP)}. IEEE, 2016.

\bibitem[Caesar et~al.(2020)Caesar, Bankiti, Lang, Vora, Liong, Xu, Krishnan, Pan, Baldan, and Beijbom]{caesar2020nuscenes}
Holger Caesar, Varun Bankiti, Alex~H Lang, Sourabh Vora, Venice~Erin Liong, Qiang Xu, Anush Krishnan, Yu Pan, Giancarlo Baldan, and Oscar Beijbom.
\newblock nuscenes: A multimodal dataset for autonomous driving.
\newblock In \emph{Proceedings of the IEEE/CVF conference on computer vision and pattern recognition}, pages 11621--11631, 2020.

\bibitem[Chen et~al.(2019)Chen, Tang, Yang, and Fu]{chen2019cooper}
Qi Chen, Sihai Tang, Qing Yang, and Song Fu.
\newblock Cooper: Cooperative perception for connected autonomous vehicles based on 3d point clouds.
\newblock In \emph{2019 IEEE 39th International Conference on Distributed Computing Systems (ICDCS)}, pages 514--524. IEEE, 2019.

\bibitem[Dosovitskiy et~al.(2017)Dosovitskiy, Ros, Codevilla, Lopez, and Koltun]{carla_2017}
Alexey Dosovitskiy, German Ros, Felipe Codevilla, Antonio Lopez, and Vladlen Koltun.
\newblock Carla: An open urban driving simulator.
\newblock In \emph{Proceedings of the 1st Annual Conference on Robot Learning}, pages 1--16. PMLR, 2017.

\bibitem[Feng et~al.(2024)Feng, Wang, Han, Zhang, and Zhu]{feng2024u2udata}
Tongtong Feng, Xin Wang, Feilin Han, Leping Zhang, and Wenwu Zhu.
\newblock U2udata: A large-scale cooperative perception dataset for swarm uavs autonomous flight.
\newblock In \emph{Proceedings of the 32nd ACM International Conference on Multimedia}, pages 7600--7608, 2024.

\bibitem[Gao et~al.(2025)Gao, Wu, Yang, Luo, Wu, Chen, Wang, Liu, Zhou, and Tu]{gao2025airv2x}
Xiangbo Gao, Yuheng Wu, Fengze Yang, Xuewen Luo, Keshu Wu, Xinghao Chen, Yuping Wang, Chenxi Liu, Yang Zhou, and Zhengzhong Tu.
\newblock Airv2x: Unified air-ground vehicle-to-everything collaboration.
\newblock \emph{arXiv preprint arXiv:2506.19283}, 2025.

\bibitem[Geiger et~al.(2012{\natexlab{a}})Geiger, Lenz, and Urtasun]{geiger2012we}
Andreas Geiger, Philip Lenz, and Raquel Urtasun.
\newblock Are we ready for autonomous driving? the kitti vision benchmark suite.
\newblock In \emph{2012 IEEE conference on computer vision and pattern recognition}, pages 3354--3361. IEEE, 2012{\natexlab{a}}.

\bibitem[Geiger et~al.(2012{\natexlab{b}})Geiger, Moosmann, Car, and Schuster]{geiger2012automatic}
Andreas Geiger, Frank Moosmann, {\"O}mer Car, and Bernhard Schuster.
\newblock Automatic camera and range sensor calibration using a single shot.
\newblock In \emph{2012 IEEE international conference on robotics and automation}, pages 3936--3943. IEEE, 2012{\natexlab{b}}.

\bibitem[Hao et~al.(2024)Hao, Fan, Dai, Zhang, Li, Wang, Yu, Yang, Yuan, and Nie]{hao2024rcooper}
Ruiyang Hao, Siqi Fan, Yingru Dai, Zhenlin Zhang, Chenxi Li, Yuntian Wang, Haibao Yu, Wenxian Yang, Jirui Yuan, and Zaiqing Nie.
\newblock Rcooper: A real-world large-scale dataset for roadside cooperative perception.
\newblock In \emph{Proceedings of the IEEE/CVF Conference on Computer Vision and Pattern Recognition}, pages 22347--22357, 2024.

\bibitem[Hu et~al.(2022)Hu, Fang, Lei, Zhong, and Chen]{hu2022where2comm}
Yue Hu, Shaoheng Fang, Zixing Lei, Yiqi Zhong, and Siheng Chen.
\newblock Where2comm: Communication-efficient collaborative perception via spatial confidence maps.
\newblock \emph{Advances in neural information processing systems}, 35:\penalty0 4874--4886, 2022.

\bibitem[Kalman(1960)]{kalman1960new}
Rudolph~Emil Kalman.
\newblock A new approach to linear filtering and prediction problems.
\newblock 1960.

\bibitem[Kingma(2014)]{kingma2014adam}
Diederik~P Kingma.
\newblock Adam: A method for stochastic optimization.
\newblock \emph{arXiv preprint arXiv:1412.6980}, 2014.

\bibitem[Koide et~al.(2021)Koide, Yokozuka, Oishi, and Banno]{koide2021voxelized}
Kenji Koide, Masashi Yokozuka, Shuji Oishi, and Atsuhiko Banno.
\newblock Voxelized gicp for fast and accurate 3d point cloud registration.
\newblock In \emph{2021 IEEE international conference on robotics and automation (ICRA)}, pages 11054--11059. IEEE, 2021.

\bibitem[Krajzewicz et~al.(2012)Krajzewicz, Erdmann, Behrisch, and Bieker]{sumo_2012}
Daniel Krajzewicz, Jakob Erdmann, Michael Behrisch, and Laura Bieker.
\newblock Recent development and applications of sumo - simulation of urban mobility.
\newblock pages 128--138, 2012.

\bibitem[Kuhn(1955)]{kuhn1955hungarian}
Harold~W Kuhn.
\newblock The hungarian method for the assignment problem.
\newblock \emph{Naval research logistics quarterly}, 2\penalty0 (1-2):\penalty0 83--97, 1955.

\bibitem[Lang et~al.(2019)Lang, Vora, Caesar, Zhou, Yang, and Beijbom]{lang2019pointpillars}
Alex~H Lang, Sourabh Vora, Holger Caesar, Lubing Zhou, Jiong Yang, and Oscar Beijbom.
\newblock Pointpillars: Fast encoders for object detection from point clouds.
\newblock In \emph{Proceedings of the IEEE/CVF conference on computer vision and pattern recognition}, pages 12697--12705, 2019.

\bibitem[Li et~al.(2020)Li, Wang, Li, Li, Wu, and Hao]{li2020sustech}
E Li, Shuaijun Wang, Chengyang Li, Dachuan Li, Xiangbin Wu, and Qi Hao.
\newblock Sustech points: A portable 3d point cloud interactive annotation platform system.
\newblock In \emph{2020 IEEE Intelligent Vehicles Symposium (IV)}, pages 1108--1115. IEEE, 2020.

\bibitem[Li et~al.(2022)Li, Ma, An, Wang, Zhong, Chen, and Feng]{li2022v2x}
Yiming Li, Dekun Ma, Ziyan An, Zixun Wang, Yiqi Zhong, Siheng Chen, and Chen Feng.
\newblock V2x-sim: Multi-agent collaborative perception dataset and benchmark for autonomous driving.
\newblock \emph{IEEE Robotics and Automation Letters}, 7\penalty0 (4):\penalty0 10914--10921, 2022.

\bibitem[Loshchilov and Hutter(2019)]{loshchilov2019decoupledweightdecayregularization}
Ilya Loshchilov and Frank Hutter.
\newblock Decoupled weight decay regularization, 2019.

\bibitem[Lu et~al.(2023)Lu, Li, Liu, Dianati, Feng, Chen, and Wang]{lu2023robust}
Yifan Lu, Quanhao Li, Baoan Liu, Mehrdad Dianati, Chen Feng, Siheng Chen, and Yanfeng Wang.
\newblock Robust collaborative 3d object detection in presence of pose errors.
\newblock In \emph{2023 IEEE International Conference on Robotics and Automation (ICRA)}, pages 4812--4818. IEEE, 2023.

\bibitem[Neubeck and Van~Gool(2006)]{neubeck2006efficient}
Alexander Neubeck and Luc Van~Gool.
\newblock Efficient non-maximum suppression.
\newblock In \emph{18th international conference on pattern recognition (ICPR'06)}, pages 850--855. IEEE, 2006.

\bibitem[Rawashdeh and Wang(2018)]{rawashdeh2018collaborative}
Zaydoun~Yahya Rawashdeh and Zheng Wang.
\newblock Collaborative automated driving: A machine learning-based method to enhance the accuracy of shared information.
\newblock In \emph{2018 21st International Conference on Intelligent Transportation Systems (ITSC)}, pages 3961--3966. IEEE, 2018.

\bibitem[Shah et~al.(2017)Shah, Dey, Lovett, and Kapoor]{shah2017airsim}
Shital Shah, Debadeepta Dey, Chris Lovett, and Ashish Kapoor.
\newblock Airsim: High-fidelity visual and physical simulation for autonomous vehicles.
\newblock In \emph{Field and service robotics: Results of the 11th international conference}, pages 621--635. Springer, 2017.

\bibitem[Shi et~al.(2020)Shi, Guo, Jiang, Wang, Shi, Wang, and Li]{shi2020pv}
Shaoshuai Shi, Chaoxu Guo, Li Jiang, Zhe Wang, Jianping Shi, Xiaogang Wang, and Hongsheng Li.
\newblock Pv-rcnn: Point-voxel feature set abstraction for 3d object detection.
\newblock In \emph{Proceedings of the IEEE/CVF conference on computer vision and pattern recognition}, pages 10529--10538, 2020.

\bibitem[Sunderraman et~al.(2024)Sunderraman, Ji, et~al.]{sunderraman2024uav3d}
Rajshekhar Sunderraman, Jonathan~Shihao Ji, et~al.
\newblock Uav3d: A large-scale 3d perception benchmark for unmanned aerial vehicles.
\newblock \emph{Advances in Neural Information Processing Systems}, 37:\penalty0 55425--55442, 2024.

\bibitem[Wang et~al.(2025)Wang, Cao, Zhong, Zhang, Yu, He, and Xu]{wang2025griffin}
Jiahao Wang, Xiangyu Cao, Jiaru Zhong, Yuner Zhang, Haibao Yu, Lei He, and Shaobing Xu.
\newblock Griffin: Aerial-ground cooperative detection and tracking dataset and benchmark.
\newblock \emph{arXiv preprint arXiv:2503.06983}, 2025.

\bibitem[Weng et~al.(2020{\natexlab{a}})Weng, Wang, Held, and Kitani]{weng20203d}
Xinshuo Weng, Jianren Wang, David Held, and Kris Kitani.
\newblock 3d multi-object tracking: A baseline and new evaluation metrics.
\newblock In \emph{2020 IEEE/RSJ International Conference on Intelligent Robots and Systems (IROS)}, pages 10359--10366. IEEE, 2020{\natexlab{a}}.

\bibitem[Weng et~al.(2020{\natexlab{b}})Weng, Wang, Held, and Kitani]{weng2020ab3dmot}
Xinshuo Weng, Jianren Wang, David Held, and Kris Kitani.
\newblock Ab3dmot: A baseline for 3d multi-object tracking and new evaluation metrics.
\newblock \emph{arXiv preprint arXiv:2008.08063}, 2020{\natexlab{b}}.

\bibitem[Xiang et~al.(2024)Xiang, Zheng, Xia, Xu, Gao, Zhou, Han, Ji, Li, Meng, et~al.]{xiang2024v2x}
Hao Xiang, Zhaoliang Zheng, Xin Xia, Runsheng Xu, Letian Gao, Zewei Zhou, Xu Han, Xinkai Ji, Mingxi Li, Zonglin Meng, et~al.
\newblock V2x-real: a largs-scale dataset for vehicle-to-everything cooperative perception.
\newblock \emph{arXiv preprint arXiv:2403.16034}, 2024.

\bibitem[Xu et~al.(2021)Xu, Guo, Han, Xia, Xiang, and Ma]{xu2021opencda}
Runsheng Xu, Yi Guo, Xu Han, Xin Xia, Hao Xiang, and Jiaqi Ma.
\newblock Opencda: an open cooperative driving automation framework integrated with co-simulation.
\newblock In \emph{2021 IEEE International Intelligent Transportation Systems Conference (ITSC)}, pages 1155--1162. IEEE, 2021.

\bibitem[Xu et~al.(2022{\natexlab{a}})Xu, Tu, Xiang, Shao, Zhou, and Ma]{xu2022cobevt}
Runsheng Xu, Zhengzhong Tu, Hao Xiang, Wei Shao, Bolei Zhou, and Jiaqi Ma.
\newblock Cobevt: Cooperative bird's eye view semantic segmentation with sparse transformers.
\newblock \emph{arXiv preprint arXiv:2207.02202}, 2022{\natexlab{a}}.

\bibitem[Xu et~al.(2022{\natexlab{b}})Xu, Xiang, Tu, Xia, Yang, and Ma]{xu2022v2x}
Runsheng Xu, Hao Xiang, Zhengzhong Tu, Xin Xia, Ming-Hsuan Yang, and Jiaqi Ma.
\newblock V2x-vit: Vehicle-to-everything cooperative perception with vision transformer.
\newblock In \emph{European Conference on Computer Vision}, pages 107--124. Springer, 2022{\natexlab{b}}.

\bibitem[Xu et~al.(2022{\natexlab{c}})Xu, Xiang, Xia, Han, Li, and Ma]{xu2022opv2v}
Runsheng Xu, Hao Xiang, Xin Xia, Xu Han, Jinlong Li, and Jiaqi Ma.
\newblock Opv2v: An open benchmark dataset and fusion pipeline for perception with vehicle-to-vehicle communication.
\newblock In \emph{2022 International Conference on Robotics and Automation (ICRA)}, pages 2583--2589, 2022{\natexlab{c}}.

\bibitem[Xu et~al.(2023)Xu, Xia, Li, Li, Zhang, Tu, Meng, Xiang, Dong, Song, et~al.]{xu2023v2v4real}
Runsheng Xu, Xin Xia, Jinlong Li, Hanzhao Li, Shuo Zhang, Zhengzhong Tu, Zonglin Meng, Hao Xiang, Xiaoyu Dong, Rui Song, et~al.
\newblock V2v4real: A real-world large-scale dataset for vehicle-to-vehicle cooperative perception.
\newblock In \emph{Proceedings of the IEEE/CVF Conference on Computer Vision and Pattern Recognition}, pages 13712--13722, 2023.

\bibitem[Yan et~al.(2018)Yan, Mao, and Li]{yan2018second}
Yan Yan, Yuxing Mao, and Bo Li.
\newblock Second: Sparsely embedded convolutional detection.
\newblock \emph{Sensors}, 18\penalty0 (10):\penalty0 3337, 2018.

\bibitem[Yin et~al.(2021)Yin, Zhou, and Krahenbuhl]{yin2021center}
Tianwei Yin, Xingyi Zhou, and Philipp Krahenbuhl.
\newblock Center-based 3d object detection and tracking.
\newblock In \emph{Proceedings of the IEEE/CVF conference on computer vision and pattern recognition}, pages 11784--11793, 2021.

\bibitem[Yu et~al.(2022)Yu, Luo, Shu, Huo, Yang, Shi, Guo, Li, Hu, Yuan, et~al.]{yu2022dair}
Haibao Yu, Yizhen Luo, Mao Shu, Yiyi Huo, Zebang Yang, Yifeng Shi, Zhenglong Guo, Hanyu Li, Xing Hu, Jirui Yuan, et~al.
\newblock Dair-v2x: A large-scale dataset for vehicle-infrastructure cooperative 3d object detection.
\newblock In \emph{Proceedings of the IEEE/CVF Conference on Computer Vision and Pattern Recognition}, pages 21361--21370, 2022.

\bibitem[Yu et~al.(2023)Yu, Yang, Ruan, Yang, Tang, Gao, Hao, Shi, Pan, Sun, et~al.]{yu2023v2x}
Haibao Yu, Wenxian Yang, Hongzhi Ruan, Zhenwei Yang, Yingjuan Tang, Xu Gao, Xin Hao, Yifeng Shi, Yifeng Pan, Ning Sun, et~al.
\newblock V2x-seq: A large-scale sequential dataset for vehicle-infrastructure cooperative perception and forecasting.
\newblock In \emph{Proceedings of the IEEE/CVF Conference on Computer Vision and Pattern Recognition}, pages 5486--5495, 2023.

\bibitem[Zhang et~al.(2024{\natexlab{a}})Zhang, Yang, Wang, Wang, Sun, and Song]{Zhang_2024_CVPR}
Jingyu Zhang, Kun Yang, Yilei Wang, Hanqi Wang, Peng Sun, and Liang Song.
\newblock Ermvp: Communication-efficient and collaboration-robust multi-vehicle perception in challenging environments.
\newblock In \emph{Proceedings of the IEEE/CVF Conference on Computer Vision and Pattern Recognition (CVPR)}, pages 12575--12584, 2024{\natexlab{a}}.

\bibitem[Zhang et~al.(2024{\natexlab{b}})Zhang, Yang, Wang, Wang, Sun, and Song]{zhang2024ermvp}
Jingyu Zhang, Kun Yang, Yilei Wang, Hanqi Wang, Peng Sun, and Liang Song.
\newblock Ermvp: Communication-efficient and collaboration-robust multi-vehicle perception in challenging environments.
\newblock In \emph{Proceedings of the IEEE/CVF Conference on Computer Vision and Pattern Recognition}, pages 12575--12584, 2024{\natexlab{b}}.

\bibitem[Zhang et~al.(2025)Zhang, Wang, Qian, Sun, Li, Jiang, Liu, and Song]{zhang2025dsrc}
Jingyu Zhang, Yilei Wang, Lang Qian, Peng Sun, Zengwen Li, Sudong Jiang, Maolin Liu, and Liang Song.
\newblock Dsrc: Learning density-insensitive and semantic-aware collaborative representation against corruptions.
\newblock In \emph{Proceedings of the AAAI Conference on Artificial Intelligence}, pages 9942--9950, 2025.

\bibitem[Zhang(2002)]{zhang2002flexible}
Zhengyou Zhang.
\newblock A flexible new technique for camera calibration.
\newblock \emph{IEEE Transactions on pattern analysis and machine intelligence}, 22\penalty0 (11):\penalty0 1330--1334, 2002.

\bibitem[Zhou et~al.(2024)Zhou, Quang, Nieto-Granda, and Loianno]{zhou2024coped}
Yang Zhou, Long Quang, Carlos Nieto-Granda, and Giuseppe Loianno.
\newblock Coped-advancing multi-robot collaborative perception: A comprehensive dataset in real-world environments.
\newblock \emph{IEEE Robotics and Automation Letters}, 9\penalty0 (7):\penalty0 6416--6423, 2024.

\bibitem[Zimmer et~al.(2024)Zimmer, Wardana, Sritharan, Zhou, Song, and Knoll]{zimmer2024tumtraf}
Walter Zimmer, Gerhard~Arya Wardana, Suren Sritharan, Xingcheng Zhou, Rui Song, and Alois~C Knoll.
\newblock Tumtraf v2x cooperative perception dataset.
\newblock In \emph{Proceedings of the IEEE/CVF Conference on Computer Vision and Pattern Recognition}, pages 22668--22677, 2024.

\end{thebibliography}
}

\clearpage
\setcounter{page}{1}
\maketitlesupplementary

\appendix
\section{Appendix}
\label{sec:appendix}

This supplementary document is organized as follows:

\begin{itemize}
    \item LiDAR appearance of different object classes are provided in~\cref{sec:sup-datasets}.
    \item More details about V2U4Real are provided in~\cref{sec:sup-dataset_detail}.
    \item Implementation details of the evaluated models are covered in~\cref{sec:sup-imp}.
    \item More experimental results are discussed in~\cref{sec:sup-detection}.
    \item Limitations and future work are discussed in~\cref{sec:limitations}.
\end{itemize}

\section{Dataset Visualization}
\label{sec:sup-datasets}
\begin{figure}[t]
\centering

\subfloat[Top view(UAV)]{
  \includegraphics[width=\columnwidth]{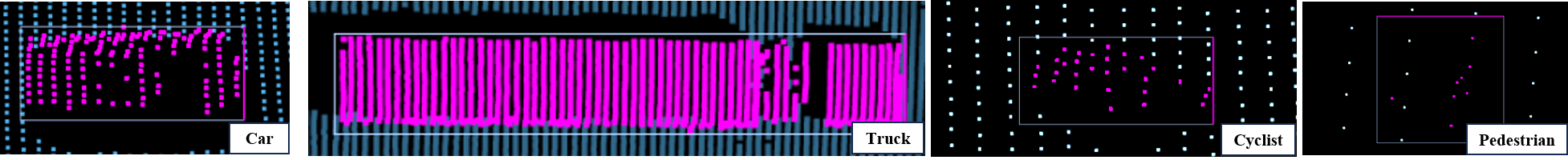}
}

\subfloat[Side view(UAV)]{
  \includegraphics[width=\columnwidth]{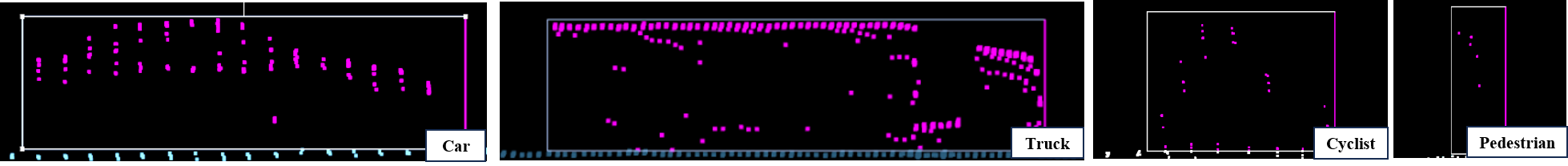}
}

\subfloat[Top view(Vehicle)]{
  \includegraphics[width=\columnwidth]{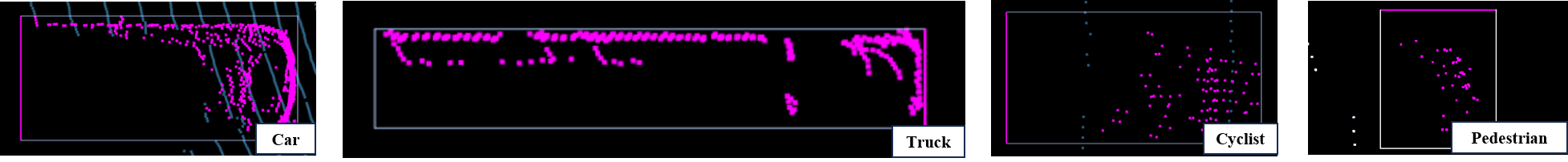}
}

\subfloat[Side view(Vehicle)]{
  \includegraphics[width=\columnwidth]{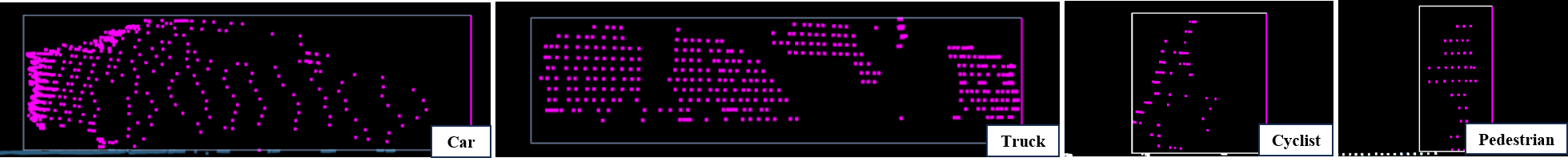}
}

\caption{\textbf{LiDAR appearance of different object classes in V2U4Real.} The first and second rows show the top and side views captured by the UAV-mounted LiDAR, while the third and fourth rows present the corresponding views from the vehicle-mounted LiDAR.}
\label{fig:object_appearance}
\end{figure}

We present the LiDAR appearance of each object class from the UAV and the ego vehicle in~\cref{fig:object_appearance}.

\section{More Details about V2U4Real}
\label{sec:sup-dataset_detail}

\subsection{Explanation of "Vehicle-to-UAV"}
The term “Vehicle-to-UAV” in our paper denotes communication connectivity rather than the direction of data flow. In our experiments, all data are transmitted from the UAV to the ego vehicle for fusion.

\subsection{Sensor Cost Reference}
We include an approximate sensor cost below for reference. 
\begin{table}[htbp]
\vspace{-4mm}
\caption{\textbf{Sensor Cost of V2U4Real.}}
\footnotesize
\centering
\renewcommand\arraystretch{0.8}
\setlength{\tabcolsep}{8.7pt}
\begin{tabular}{c|c|c}
\toprule
 Sensor & Model& Cost (\$) \\
\midrule
 LiDAR & OS-128 / RS-128 / M1-Plus & 12K / 9K / 3K \\
 Camera & HikRobot MV-CA023-10GC & 700\\
 GPS/IMU & Quectel / Sigma100 / DETA100 & 35 / 1K / 200\\
\bottomrule
\end{tabular}
\vspace{-4mm}
\label{tab:sup_sensor_cost}
\end{table}

\subsection{Time Synchronization}
\label{sec:sup-time_sync}
For all cooperative perception datasets, achieving temporal synchronization between sensors is crucial. To ensure a unified temporal reference, all onboard computing units are first synchronised to the GPS time standard. Subsequently, hardware-level synchronisation of LiDAR and camera sensors is accomplished through the combined use of the Precision Time Protocol (PTP) and Pulse Per Second (PPS) signals. Afterward, the temporally closest LiDAR frames from the ego vehicle and UAV are paired, and the corresponding camera data are temporally aligned with each LiDAR frame to construct coherent multi-modal data samples. The inter-sensor temporal discrepancy across the two agents is maintained within 20 milliseconds for each recorded sample.

\begin{figure}[!htb]
\hfil
\subfloat{%
  \includegraphics[width=1\columnwidth]{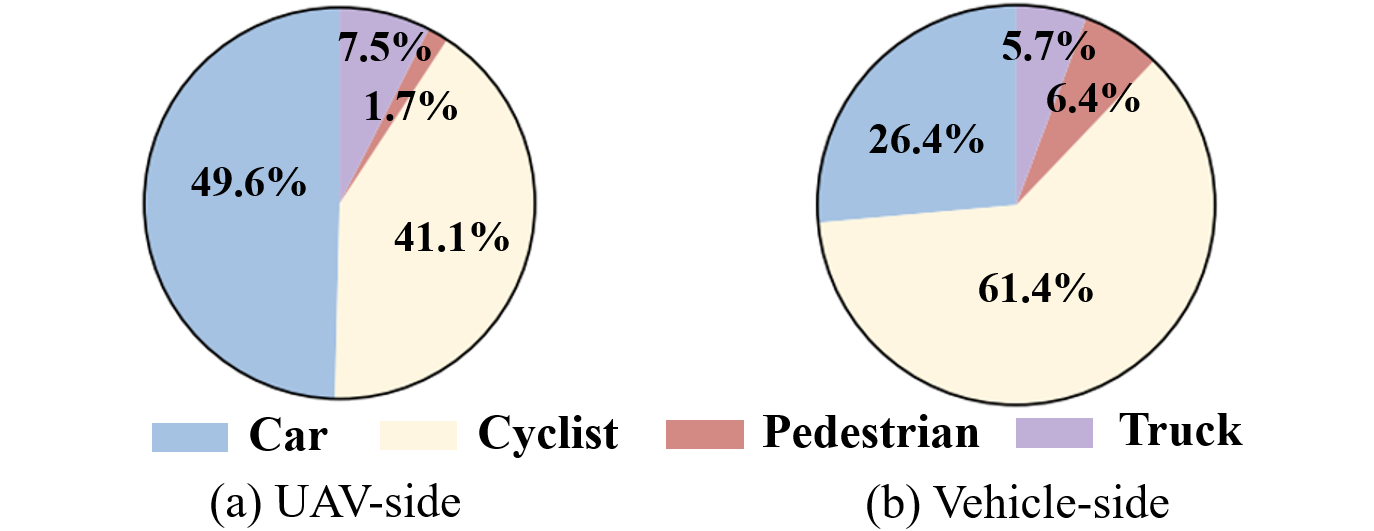}%
}
\caption{\textbf{Category distributions of V2U4Real.}}
\label{fig:category_distributions}
\end{figure}

\subsection{Dataset Statistics}
\label{sec:sup-spilt_statistics}
To enable comprehensive evaluation and fair comparison across various cooperative perception tasks, V2U4Real dataset is split into train/val/test sets at the sequence level. This design prevents frame-level overlap and ensures that models evaluated on the validation and test sets are not exposed to temporally adjacent frames from the training set. The samples are evenly distributed across the three splits to maintain a balanced representation of scene types, object densities, and motion patterns. Tab.~\ref{tab:dataset_statistics} summarizes the frame-level statistics for each split, including the number of data frames and total 3D annotations. The test split is deliberately made more challenging, with greater scene diversity and denser traffic interactions, allowing for a robust assessment of model generalization in real-world cooperative perception scenarios.

\newcolumntype{C}{>{\centering\arraybackslash}X}

\begin{table*}[!t]
\footnotesize
\centering
\caption{\textbf{Dataset statistics of V2U4Real.}}
\label{tab:dataset_statistics}

\setlength{\tabcolsep}{3pt}
\begin{tabularx}{\textwidth}{c|c|c|C|C|C|C|C|C}
\toprule
Split & Sequence & Scene & 
Frames & 
Vehicles & Cyclists & Pedestrians &
Track IDs & Duration (s)\\
\midrule
		
\multirow{24}{*}{Train}
& 2025-07-17-16-12-1 & Urban  & 1200&7140  & 1036 &0  & 112  & 30.0 \\
& 2025-07-17-16-12-2 & Urban & 1244&9255  &1386  &0   &138   & 31.1\\
& 2025-07-17-16-12-3 & Urban & 1116& 13865 &1116 &0  & 145&  27.9\\
& 2025-07-17-16-12-4 & Urban  &1208 & 8890 &1551 &0  & 129&  30.2 \\	
& 2025-07-17-16-35-3 & Urban  &3112&10175  &18342 & 2730 &241 & 77.8 \\
& 2025-07-17-16-50-1 & Urban  &1528&7247  &14733 &1953  &207 & 38.2\\
& 2025-07-17-16-50-2 & Urban  &1532&5748  &25348 &1299  &324 & 38.3\\
& 2025-07-17-16-50-3 & Urban  &1748& 4526 &15532&616& 171& 43.7\\
& 2025-07-17-17-07-2 & Campus  &1744&1986  & 8592&0  &61 & 43.6\\
& 2025-07-17-17-07-6 & Campus  &1588&13805  &19806 &1291  &242 & 39.7\\
& 2025-07-17-17-42-1 & Campus  &1264&8743  &12596 &3695  &154 & 31.6\\
& 2025-07-17-17-42-2 & Campus  &1244&1566  &6077 &1728  & 88& 31.1\\
& 2025-07-17-17-42-3 & Campus  &884& 2457 & 3405&1930  & 83& 22.1\\
& 2025-07-17-17-42-4 & Campus  &1120&4673  &8144 & 2791 &127 & 28.0\\
& 2025-07-17-17-42-5 & Campus  &732& 1097 &12328 &1275  & 153& 18.3\\
& 2025-07-18-12-10-1 & Campus  &1224&1387  & 23034& 1670 & 266& 30.6\\
& 2025-07-18-12-10-2 & Campus  &1284&1154  & 16701& 3340 & 226& 32.1\\
& 2025-07-18-12-10-3 & Campus  &1768&1228  & 29856& 655 & 360& 44.2\\   
& 2025-07-18-12-37-1 & Rural  &1240&8141  &2379 &311 & 74& 31\\
& 2025-07-18-12-37-2 & Rural  &1936&15650  &6899 &1395  & 185& 48.4\\
& 2025-07-18-12-37-3 & Rural  &1400&6209  &1800&0& 76& 35\\
& 2025-07-18-12-53-1 & Rural  &1212&2162  & 1622&130  &26& 30.3\\
& 2025-07-18-12-53-2 & Rural  &876&2021& 875&0  &21 & 21.9\\
& 2025-07-18-12-53-3 & Rural  &1340& 4735 & 1027& 0 & 42& 33.5\\
\midrule
			
\multirow{9}{*}{Val}
& 2025-07-17-16-12-7 & Urban &364& 2460 & 364 &0  & 65&9.1\\
& 2025-07-17-16-12-8 & Urban &544& 3686 & 375 & 0 & 56&13.6\\
& 2025-07-17-16-35-2 & Urban &784&932  &19775  &116  & 309&19.6\\
& 2025-07-17-16-50-6 & Urban &2128&3880  &27245  &3398  &342 &53.2\\
& 2025-07-17-17-07-1 & Campus &1524& 7355 & 10731 &1343  &134 &38.1\\
& 2025-07-17-17-42-7 & Campus &364& 552 &2738  &121  & 67&9.1\\
& 2025-07-18-12-10-4 & Campus &1280& 1843 & 15853 & 663 & 228&32.0\\
& 2025-07-18-12-37-5 & Rural &1400&3160  & 1682 & 0 & 51&35.0\\
& 2025-07-18-12-53-4 & Rural &1508& 8417 &2471  &8  &185 &37.7\\
\midrule
\multirow{9}{*}{Test}
& 2025-07-17-16-12-6 & Urban &1036& 7453 & 1016 &0  & 135& 25.9\\
& 2025-07-17-16-35-1 & Urban &1096&8431  &6063  &410  & 166& 27.4\\
& 2025-07-17-16-50-4 & Urban &1460&1094  &26474  &149  &303 & 36.5\\
& 2025-07-17-16-50-5 & Urban &1156&867 &16620  & 772 & 327& 28.9\\
& 2025-07-17-17-07-3 & Campus &1108&1108  & 21746 &1110  &240 & 27.7\\
& 2025-07-17-17-42-6 & Campus &700& 802 &12820  & 0 & 140& 17.5\\
& 2025-07-17-17-42-8 & Campus &476&1273 &4859&826&96 & 11.9\\
& 2025-07-18-12-10-5 & Rural &1896&15997  & 16860 & 4962 & 413& 47.4\\
& 2025-07-18-12-53-5 & Rural &1616&12559  &1654  &533  &126 & 40.4\\
\bottomrule
\end{tabularx}
\end{table*}

\subsection{Annotation Process and Labeler Details}
We use a two-round annotation pipeline. (1) \textbf{44} professional annotators are responsible for specific scenarios. (2) \textbf{11} quality-control annotators refine annotations across four scenarios.

\begin{figure}[!htb]
\hfil
\subfloat{%
  \includegraphics[width=1\columnwidth]{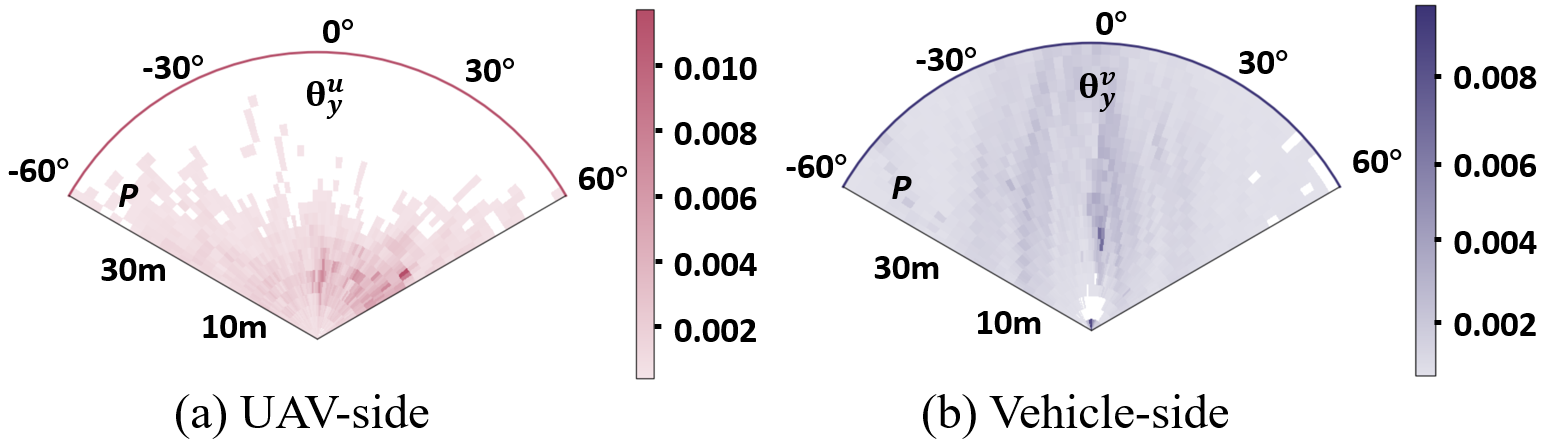}%
}
\caption{\textbf{Target bounding box distributions of V2U4Real.}}
\label{fig:bounding_box_distributions}
\end{figure}

\begin{table*}[t]  
\footnotesize
\centering
\addtolength{\tabcolsep}{-2pt}
\caption{\textbf{Single-agent 3D object detection benchmarks on V2U4Real val set.} The vehicle category includes car and truck.}
\newcolumntype{C}{>{\centering\arraybackslash}X}  
\begin{tabularx}{\textwidth}{c|c|C C|C C|C C} 
\toprule
\multirow{2}{*}{LiDAR Type} & \multirow{2}{*}{Method}  & \multicolumn{2}{c|}{Vehicle (3D AP@IoU)} & \multicolumn{2}{c|}{Cyclist (3D AP@IoU)} & \multicolumn{2}{c}{Pedestrian (3D AP@IoU)} \\ 
\cmidrule(lr){3-8}
& & 0.5 & 0.7 & 0.25 & 0.5 & 0.25 & 0.5 \\
\midrule
\multirow{4}{*}{Ruby-128}
& PointPillars~\cite{lang2019pointpillars} & 60.57 & 29.57 & 59.62 & 45.30 & 38.08 & 27.65 \\
& SECOND~\cite{yan2018second} & 56.61 & 32.01 & 52.73 & 42.29 & 48.15 & 36.97 \\
& CenterPoint~\cite{yin2021center} & 58.24 & 26.93 & 56.39 & 42.56 & \textbf{58.79} & \textbf{43.64} \\
& PV-RCNN~\cite{shi2020pv} & \textbf{65.91} & \textbf{41.99} & \textbf{67.11} & \textbf{53.35} & 45.06 & 37.92 \\
\midrule
\multirow{4}{*}{M1-Plus}
& PointPillars~\cite{lang2019pointpillars} & 41.37 & 20.81 & 53.28 & 41.40 & 15.06 & 9.09 \\
& SECOND~\cite{yan2018second} & 34.73 & 18.04 & 54.67 & 42.09 & 24.20 & 16.03 \\
& CenterPoint~\cite{yin2021center} & \textbf{41.84} & 19.75 & \textbf{60.62} & 42.54 & \textbf{24.81} & \textbf{16.35} \\
& PV-RCNN~\cite{shi2020pv} & 41.38 & \textbf{22.27} & 59.12 & \textbf{45.89} & 12.57 & 7.58 \\
\bottomrule
\end{tabularx}
\label{tab:sup_single_detection}
\end{table*}

\newcolumntype{C}{>{\centering\arraybackslash}X} 

\begin{table*}[!t]
\footnotesize
\centering
\caption{\textbf{Cooperative 3D object detection benchmarks for vehicle category on V2U4Real test set.} Sync. means synchronous setup ignoring communication delays. Async. implies asynchronous setup with a $(0, 1000]$\,ms delay.}
\setlength{\tabcolsep}{3pt}
\begin{tabularx}{\textwidth}{c|c|c|CCC|CCC|C}
\toprule
\multirow{3}{*}{Paradigm} & \multirow{3}{*}{Method} & \multirow{3}{*}{Year} & 
\multicolumn{3}{c|}{Sync. (3D AP@IoU = 0.5 / 0.7)} & 
\multicolumn{3}{c|}{Async. (3D AP@IoU = 0.5 / 0.7)} & 
\multicolumn{1}{c}{AM} \\
\cmidrule(r){4-9}
& & &Overall & 0–50m & 50–100m & Overall & 0–50m & 50–100m & (MB)\\
\midrule

\multirow{9}{*}{Vehicle to UAV}
& Early Fusion & - & \textbf{45.44/20.76} & \textbf{49.59/22.38} & \textbf{20.56/13.16} & 24.63/9.63 & 26.29/9.71 & 16.53/11.01 & 3.18 \\
& Late Fusion & - & 36.76/13.20 & 39.89/14.06 & 11.53/8.29 & 23.48/8.14 & 24.62/8.38 & 10.22/7.37 & 0.009 \\
& Where2comm~\cite{hu2022where2comm} & 2022 & 33.14/12.14 & 36.31/12.94 & 16.69/12.45 & 30.68/10.81 & 33.50/11.26 & 16.58/12.44 & 0.65 \\
& V2X-ViT~\cite{xu2022v2x}& 2022 & 29.69/11.65 & 33.24/12.63 & 10.92/7.93 & 29.72/11.62 & 33.23/12.76 & 10.87/7.80 &  0.65 \\
& AttFuse~\cite{xu2022opv2v} & 2022 & 38.75/17.22 & 43.17/18.95 & 17.57/12.23 & 29.76/11.65 & 33.00/11.87 & 17.06/11.87 & 0.65 \\
& CoBEVT~\cite{xu2022cobevt} & 2022 & 31.52/14.45 & 36.26/16.59 & 13.85/8.72 & 25.27/9.77 & 28.50/10.53 & 13.41/8.41 & 0.65 \\
& CoAlign~\cite{lu2023robust}& 2023  & 41.78/18.79 & 46.06/20.17 & 17.75/13.42 & \textbf{36.41/15.58} & \textbf{39.88/16.98} & \textbf{17.15/13.04} & 0.65 \\
& ERMVP~\cite{Zhang_2024_CVPR} & 2024 & 36.48/14.28 & 40.97/15.76 & 13.22/8.21 & 24.46/8.07 & 27.29/8.44 & 10.87/7.51 & 0.65 \\
& DSRC~\cite{zhang2025dsrc} & 2025 & 41.88/17.27 & 45.73/18.48 & 18.32/12.88 & 35.38/13.38 & 38.27/13.91 & 16.75/12.63 & 0.65 \\
\midrule
Vehicle only & No Fusion & - & 28.25/10.05 & 31.13/12.05 & 10.11/6.01 & 28.25/10.05 & 31.13/12.05 & 10.11/6.01 & 0 \\
\midrule
UAV only & No Fusion &- & 31.47/11.06 & 33.21/12.42 & 12.52/8.13 & 31.47/11.06 &33.21/12.42& 12.52/8.13 & 0 \\
\bottomrule
\end{tabularx}
\label{tab:sup_coperception}
\end{table*}

\subsection{Category Distributions}
Fig.~\ref{fig:category_distributions} shows the distribution of object types in V2U4Real across different agents. On the UAV side, Car constitutes the majority at 49.6\%, followed by Cyclist at 41.1\% and Trucks at 7.5\%, while Pedestrian makes up only 1.7\%. On the vehicle side, Cyclist account for the largest proportion at 61.4\%, followed by Car at 26.4\%, Trucks at 5.7\%, and Pedestrian at 6.4\%.

\subsection{Target Bounding Box Distributions}
\label{sec:sup-bounding_box_distributions}
Variations in sensing altitude and observation geometry lead to markedly different BEV distribution patterns across agents. As shown in ~\cref{fig:bounding_box_distributions}, UAV-side and vehicle-side annotations exhibit distinct spatial and angular characteristics. On the UAV side, targets occupy a broad and relatively sparse BEV region, with higher counts around the central downward viewing direction. This arises from the UAV’s elevated bird’s-eye viewpoint: ground objects are projected across a wide radial range, resulting in larger dispersion and higher variance in BEV distances. Moreover, the extended visibility from aerial perspectives leads to targets appearing within a wide horizontal angular span $\theta_{y}^{u} \in [-60^{\circ}, 60^{\circ}]$, further contributing to the irregular and non-uniform distribution. In contrast, the vehicle-side distribution is significantly more compact. Due to the vehicle’s low and stable sensing height, detections are concentrated in a narrower, forward-looking region. Targets mainly appear near the frontal direction, with shorter BEV distances and much smaller angular variation $\theta_{y}^{v} \in [-10^{\circ}, 10^{\circ}]$. The limited elevation also reduces the likelihood of observing distant objects, yielding lower variance and a higher density of detections around the ego origin.

\section{Implementation Details}
\label{sec:sup-imp}
\subsection{Detection Model Settings}
We provide additional implementation details of the baseline methods used in our experiments to facilitate accurate reproduction of the benchmark results. For the single-agent 3D object detection task, we set the maximum number of points per voxel to 32 and the maximum number of voxels to 160,000. The models are trained for 40 epochs with an initial learning rate of 0.002. Additional optimization and preprocessing configurations are summarised in~\cref{tab:sup_imp_details_single}. For the cooperative 3D object detection task, we similarly set the maximum points per voxel to 32, while the maximum voxel count is reduced to 32,000 due to the multi-agent fusion input. The same training schedule with single-agent object detection task and a voxel size of [0.4, 0.4, 8] is adopted. The remaining settings can be found in~\cref{tab:sup_imp_details_cooperative}.

\subsection{Asynchronous Communication Modeling}
Following ~\cite{xu2023v2v4real, xu2022v2x}, we simulated asynchronous communication by temporally misaligning sensor frames and modeling delay as the sum of system overhead, transmission time, and backbone computation latency.

\section{More Experimental Results}
\label{sec:sup-detection}
\subsection{Results on Ruby-128 and M1-Plus LiDAR}
\cref{tab:sup_single_detection} shows the single-agent 3D object detection performance on the V2U4Real val set using two heterogeneous LiDAR sensors: Ruby-128 and M1-Plus. We evaluate four representative baseline detectors, including PointPillars~\cite{lang2019pointpillars}, SECOND~\cite{yan2018second}, CenterPoint~\cite{yin2021center}, and PV-RCNN~\cite{shi2020pv}. These results reveal the varying robustness of different 3D detectors under heterogeneous LiDAR sensors.

\begin{table}[t]
\centering
\footnotesize
\caption{\textbf{Implementation details of single-agent 3D object detection task.}}
\label{tab:sup_imp_details_single}
\setlength{\tabcolsep}{3pt}
\begin{tabular}{l|c|c|c|c|c}
\toprule
Method & Epoch & Batch & LR & LR Scheduler & Voxel Size \\
\midrule
PointPillar\cite{lang2019pointpillars} & 40  & 4 & 0.002 & OneCycle & $[0.2,0.2,8]$ \\
SECOND\cite{yan2018second}             & 40 & 4 & 0.002 & OneCycle & $[0.1,0.1,0.2]$ \\
CenterPoint\cite{yin2021center}        & 40  & 4 & 0.002 &  OneCycle & $[0.1,0.1,0.2]$ \\
PV-RCNN\cite{shi2020pv}                & 40  & 4 & 0.002 & OneCycle & $[0.1,0.1,0.2]$ \\
\bottomrule
\end{tabular}
\end{table}

\subsection{Results on the Test Set}
\cref{tab:sup_coperception} shows the cooperative 3D object detection results for the vehicle category on the V2U4Real test set, evaluated under both synchronous and asynchronous settings. Compared with the validation results reported in the main paper (see~\cref{tab:exps_coperception}), the test split presents significantly greater challenges, including more frequent pose variations, denser traffic, and more complex occlusion patterns. Under the synchronous setting, Early Fusion achieves the best performance, indicating that existing intermediate fusion methods still have substantial room for improvement when dealing with more complex real-world scenarios. However, under the asynchronous setting, Early Fusion fails to effectively handle communication delays (consistent with the observations in~\cref{tab:exps_coperception}), leading to a significant performance drop, while CoAlign~\cite{lu2023robust} achieves the best performance.

\subsection{Visualization of Cooperative 3D Detection}

\begin{figure*}[!ht]
\centering
\footnotesize

\begin{tabular}{ccc}
\includegraphics[width=0.3\linewidth,height=0.16\linewidth]{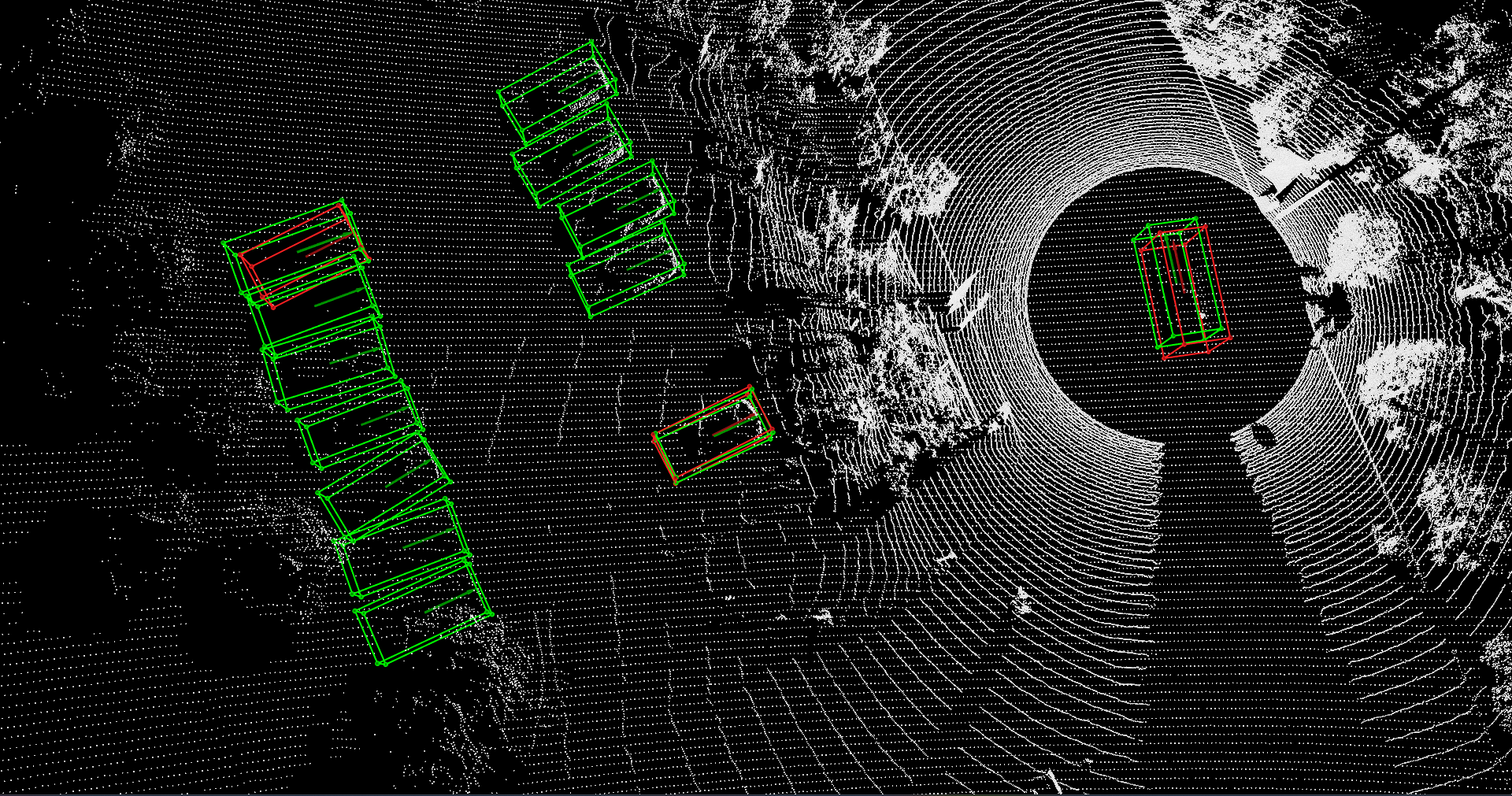} &
\includegraphics[width=0.3\linewidth,height=0.16\linewidth]{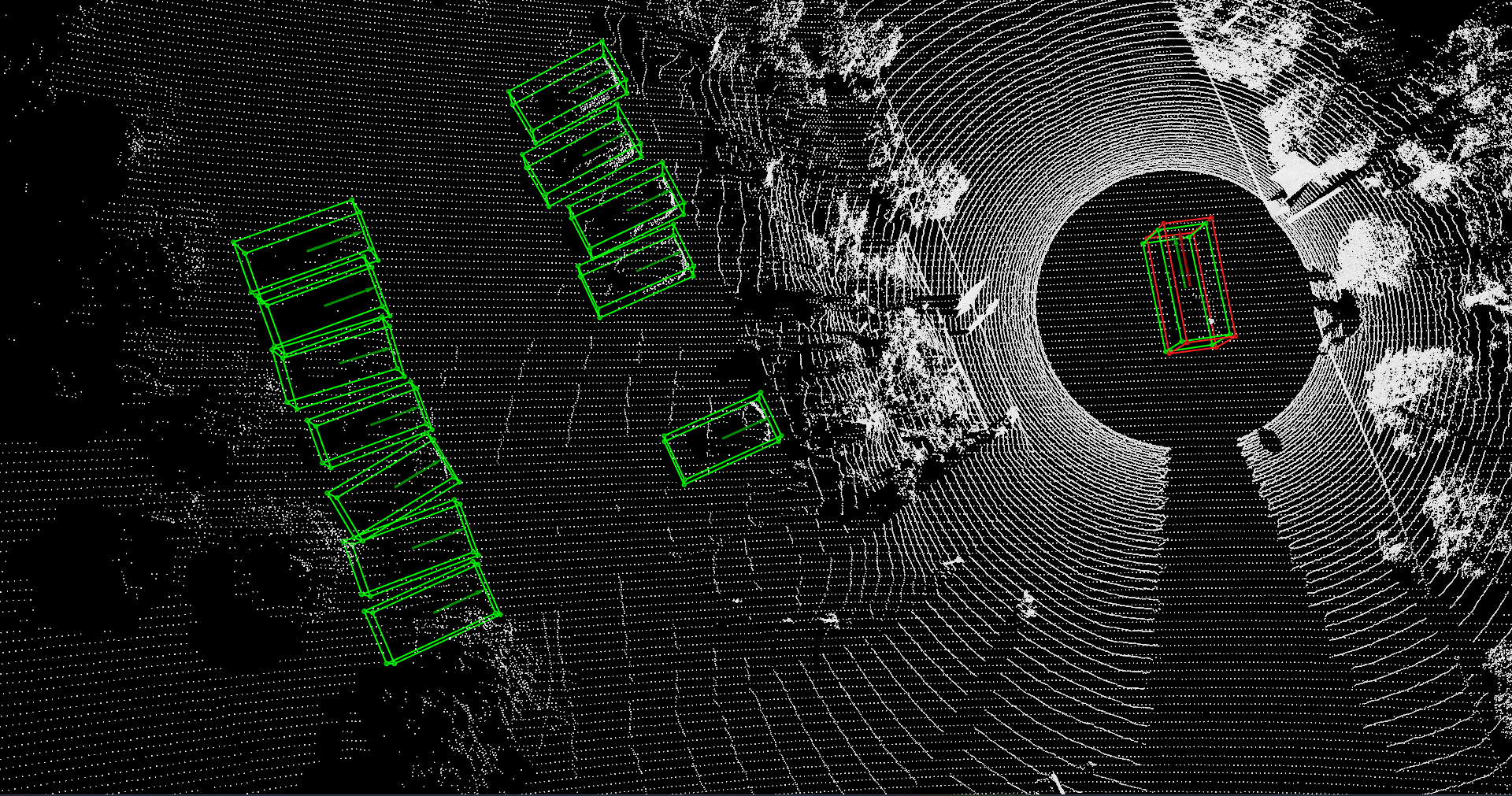} &
\includegraphics[width=0.3\linewidth,height=0.16\linewidth]{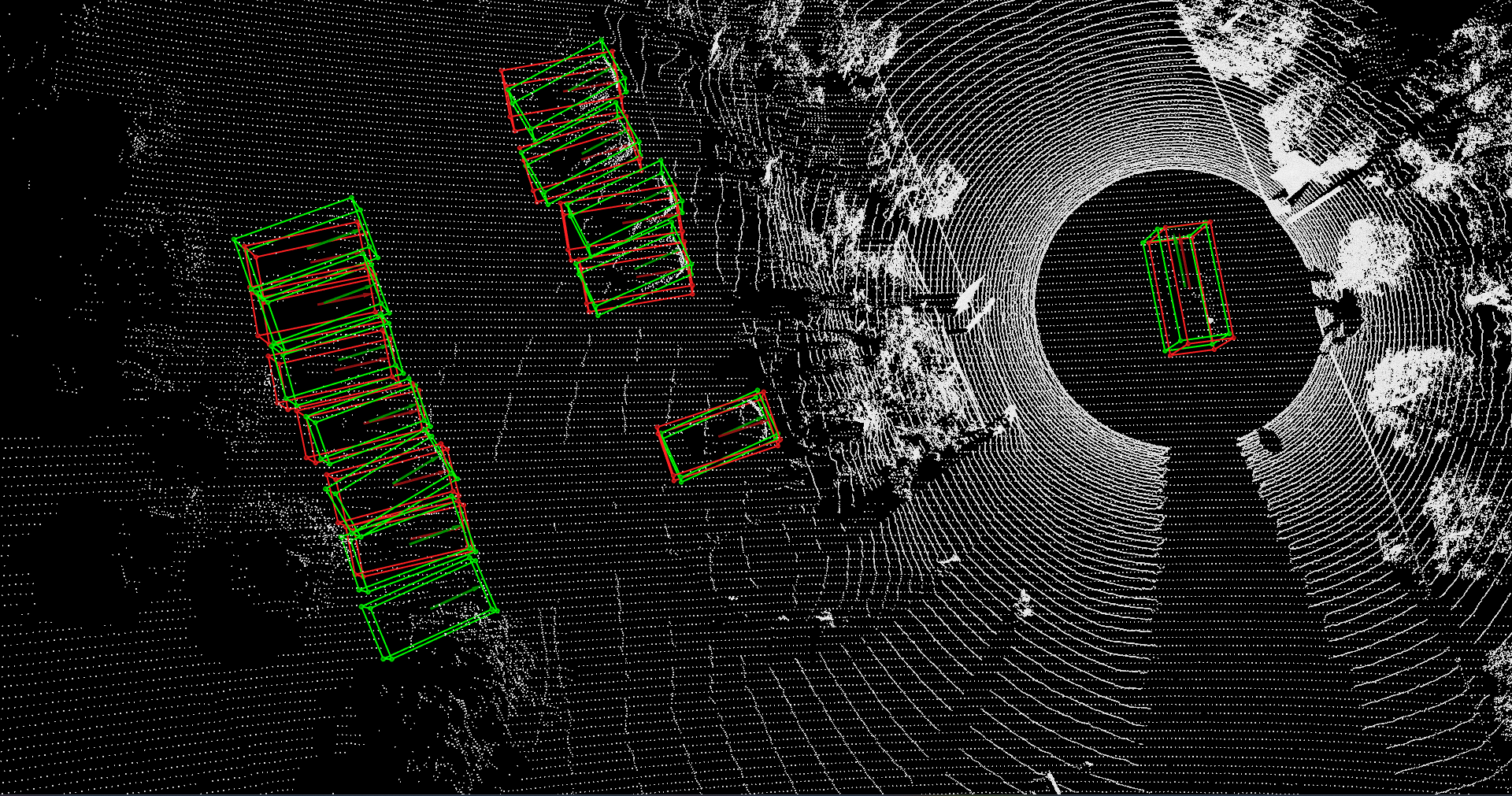} \\
Early Fusion & Late Fusion &  Where2comm~\cite{hu2022where2comm}\\
\includegraphics[width=0.3\linewidth,height=0.17\linewidth]{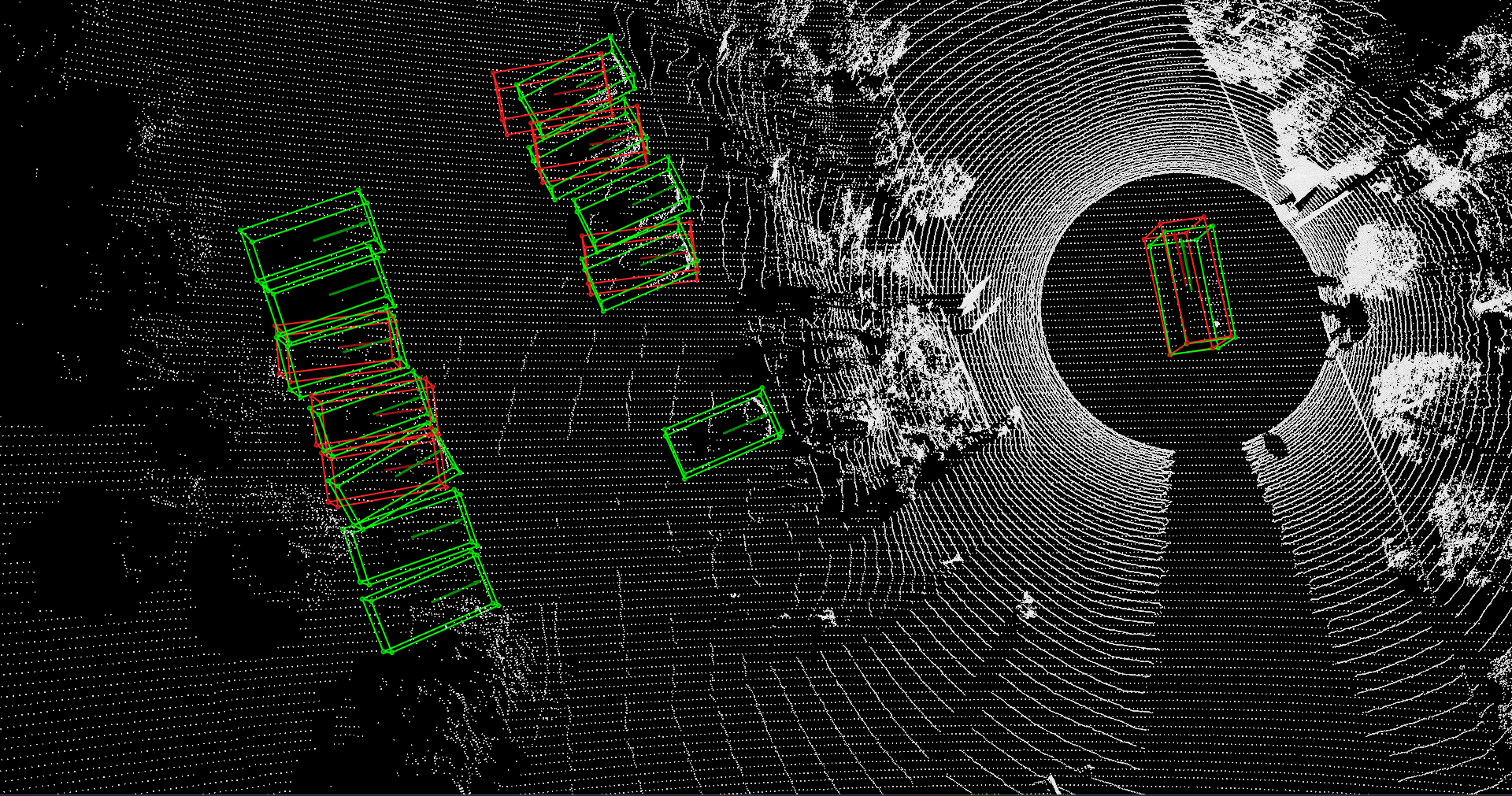} &
\includegraphics[width=0.3\linewidth,height=0.17\linewidth]{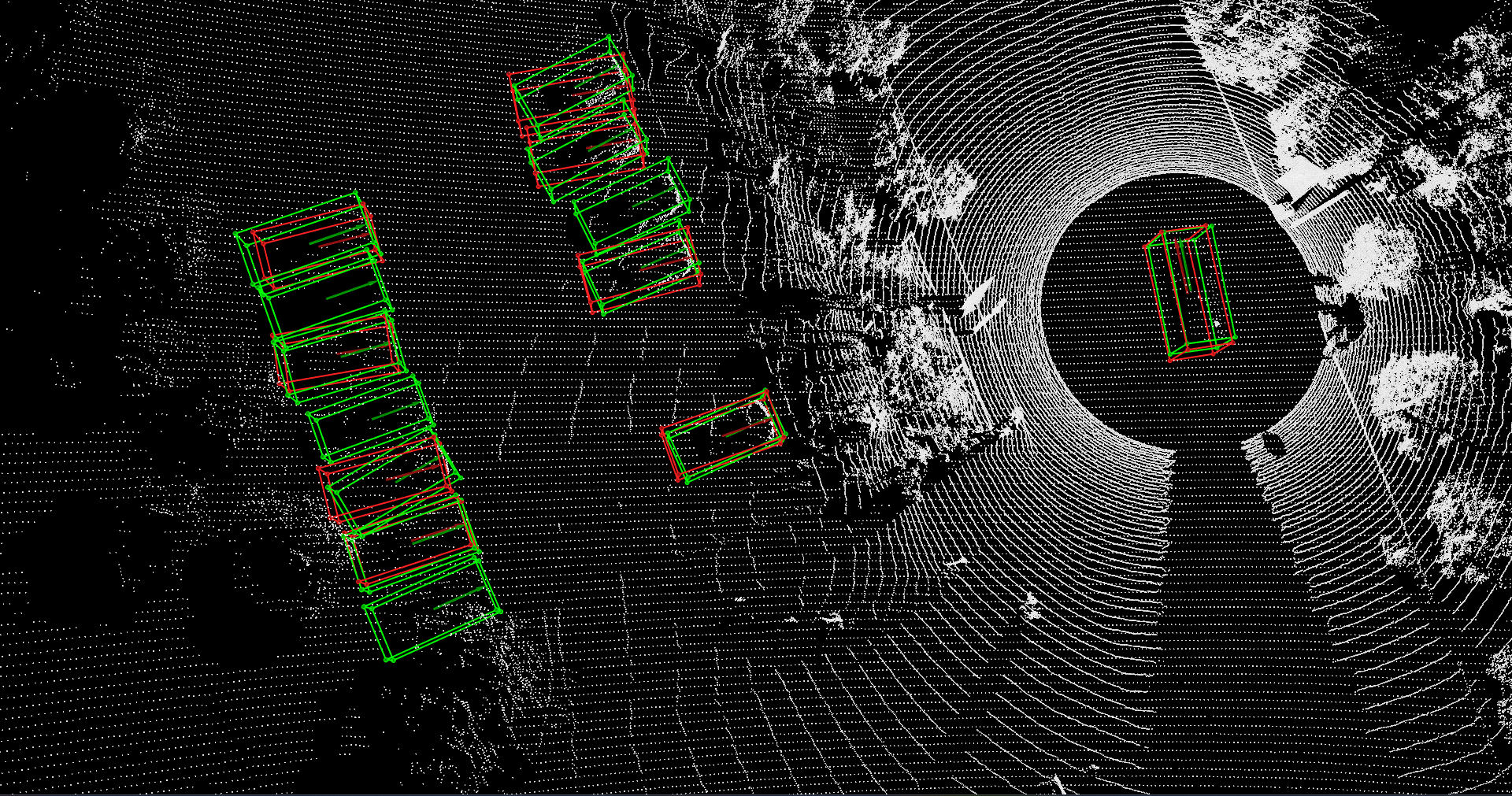} &
\includegraphics[width=0.3\linewidth,height=0.17\linewidth]{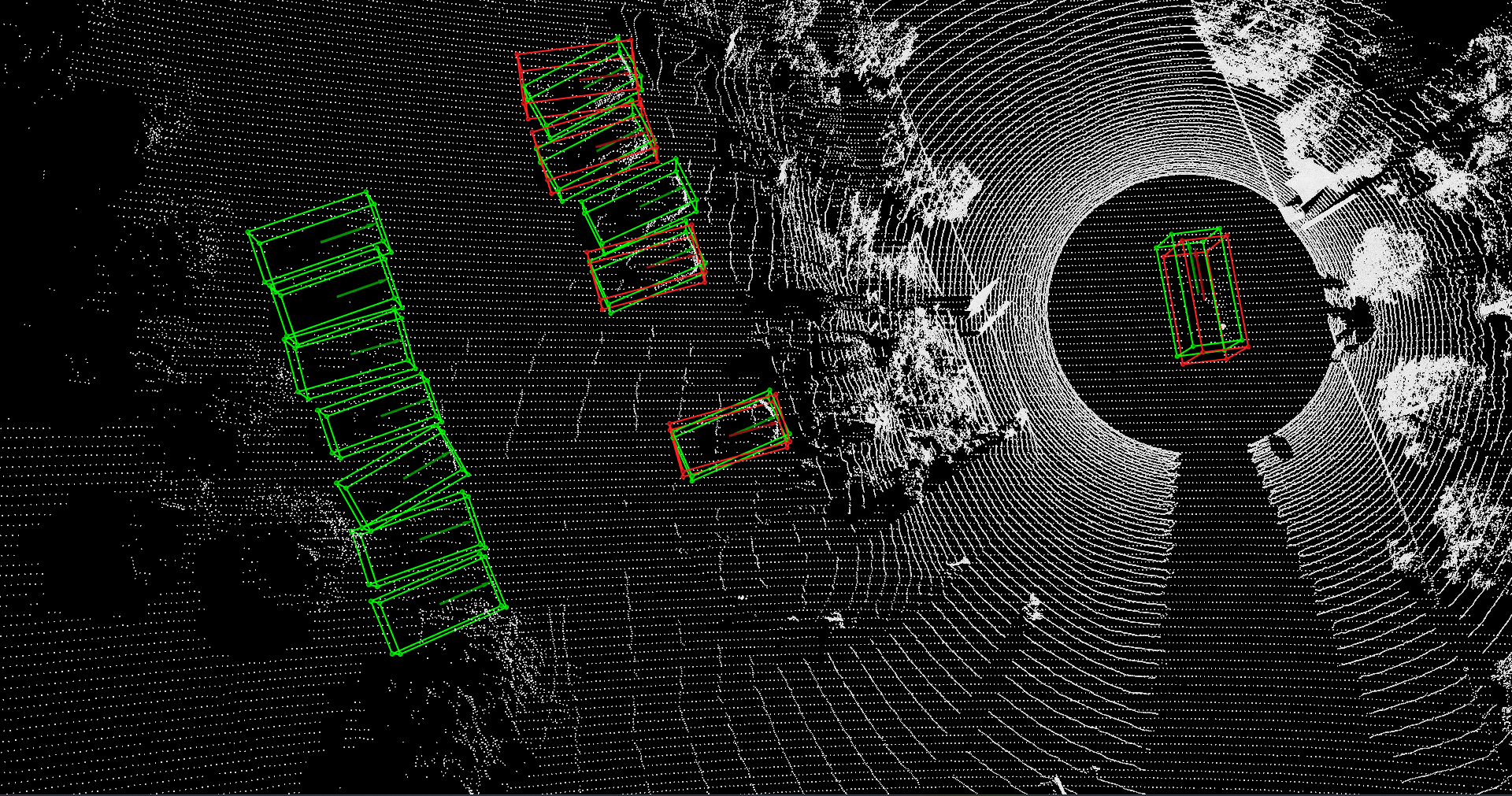} \\
AttFuse~\cite{xu2022opv2v} & CoAlign~\cite{lu2023robust} & DSRC~\cite{zhang2025dsrc} \\
\end{tabular}

\vspace{-3mm}
\caption{\textbf{Visualization of 3D cooperative detection results in scene 1.} \textcolor{green}{Green bounding boxes} indicate ground-truth annotations. \textcolor{red}{Red bounding boxes} indicate the model's predictions.}
\label{fig:vis_1}
\vspace{-4mm}
\end{figure*}

\begin{figure*}[!ht]
\centering
\footnotesize
\begin{tabular}{ccc}
\includegraphics[width=0.3\linewidth,height=0.16\linewidth]{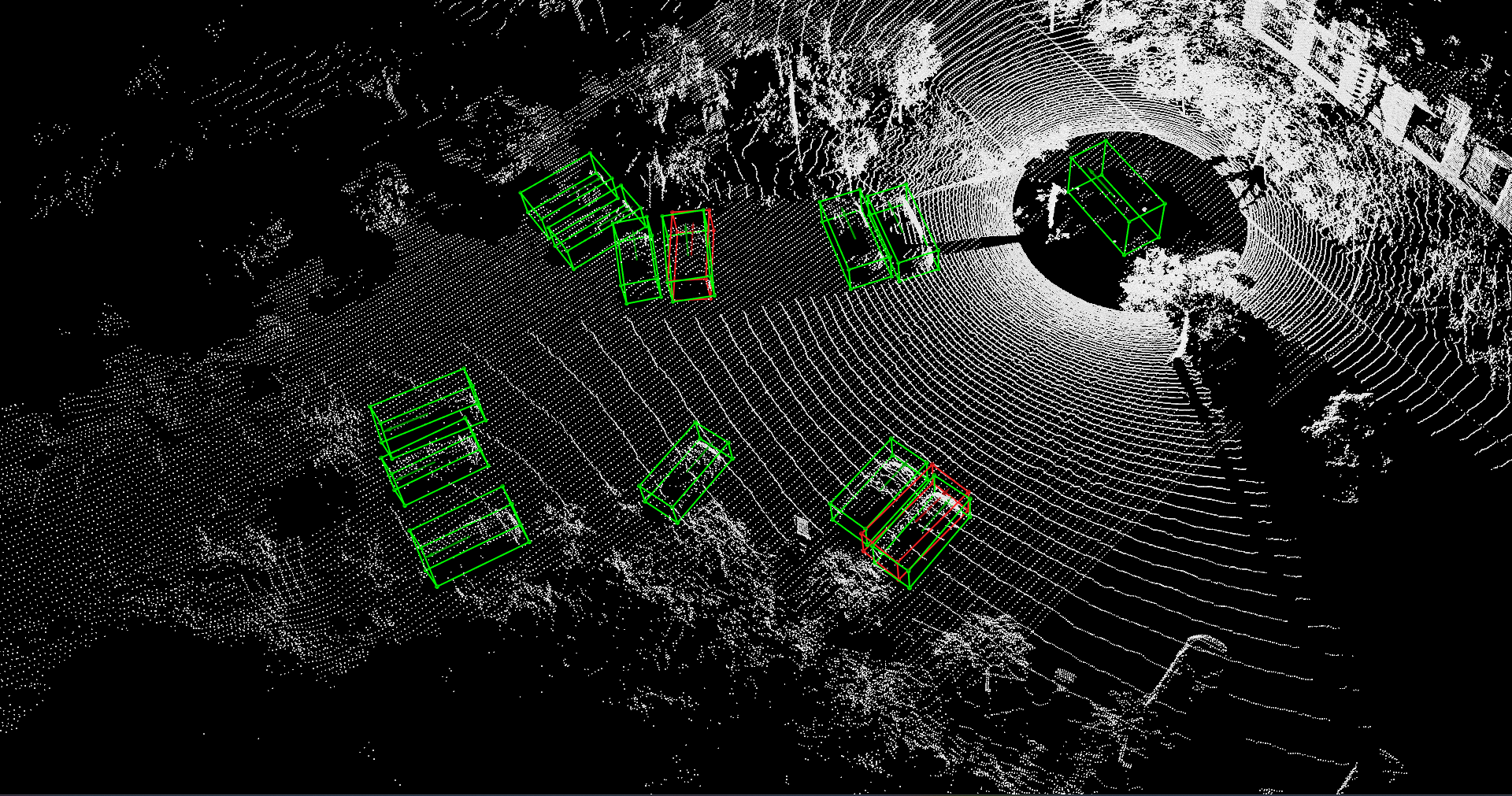} &
\includegraphics[width=0.3\linewidth,height=0.16\linewidth]{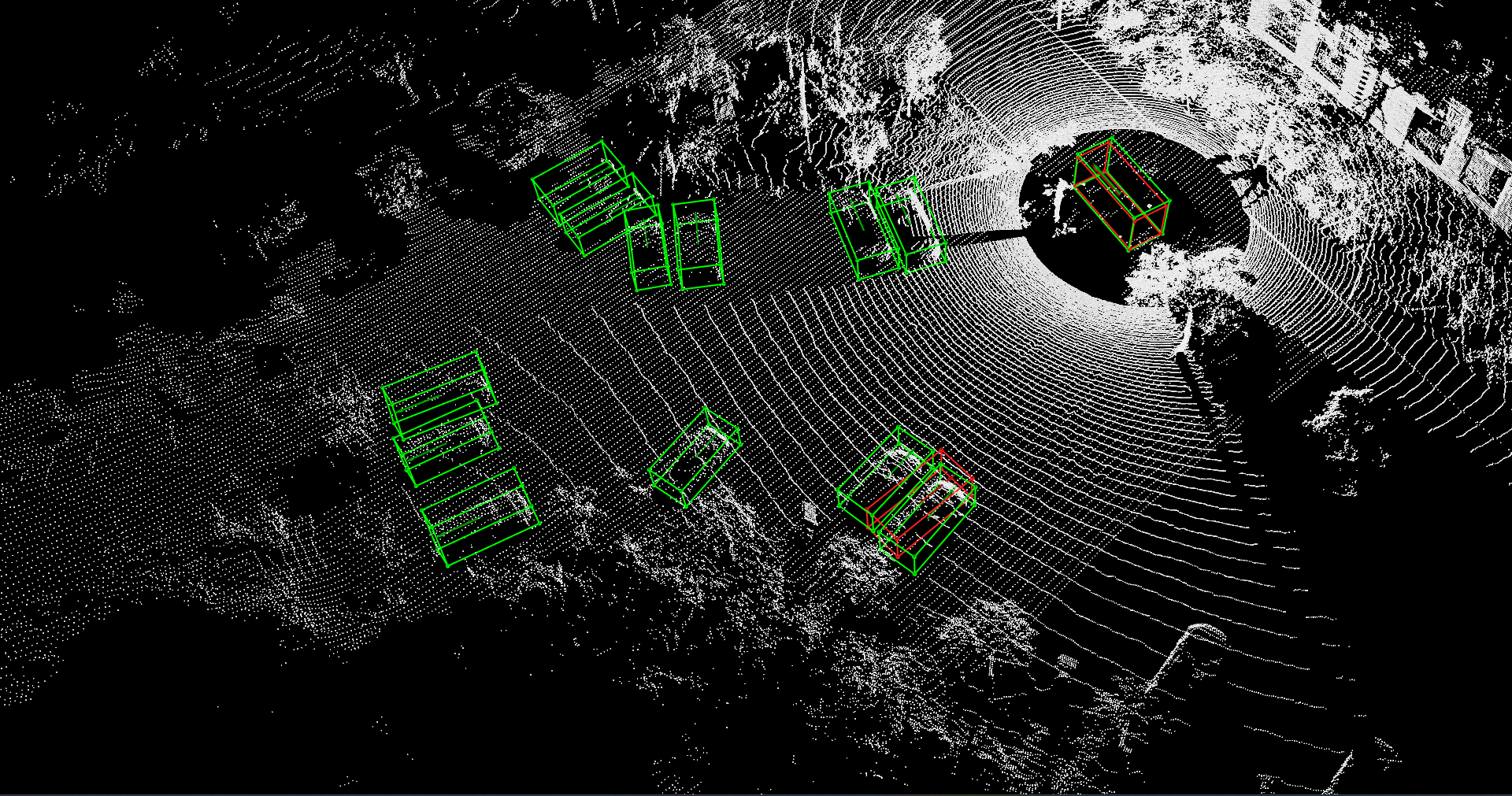} &
\includegraphics[width=0.3\linewidth,height=0.16\linewidth]{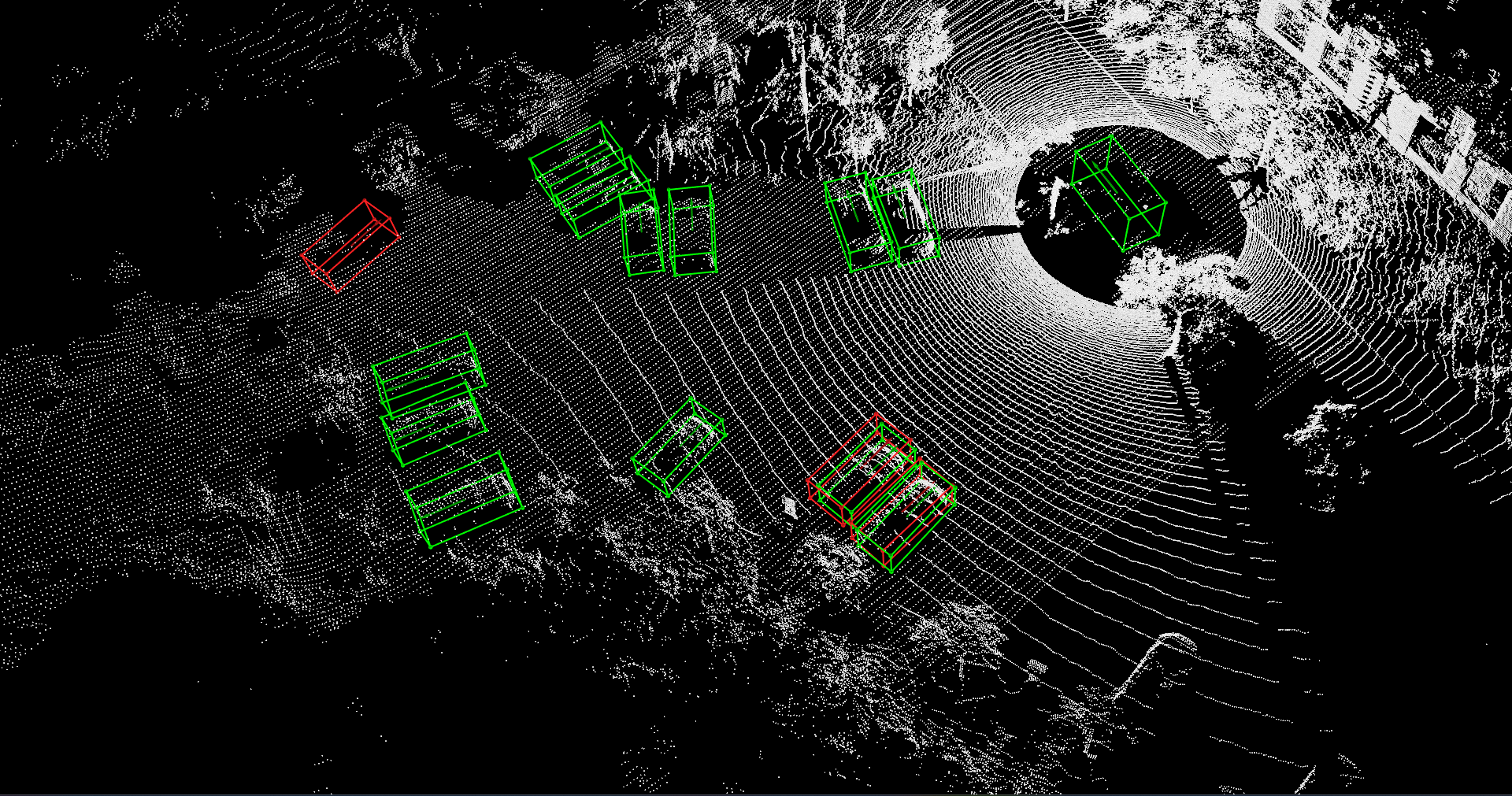} \\
Early Fusion & Late Fusion &  Where2comm~\cite{hu2022where2comm}\\
\includegraphics[width=0.3\linewidth,height=0.17\linewidth]{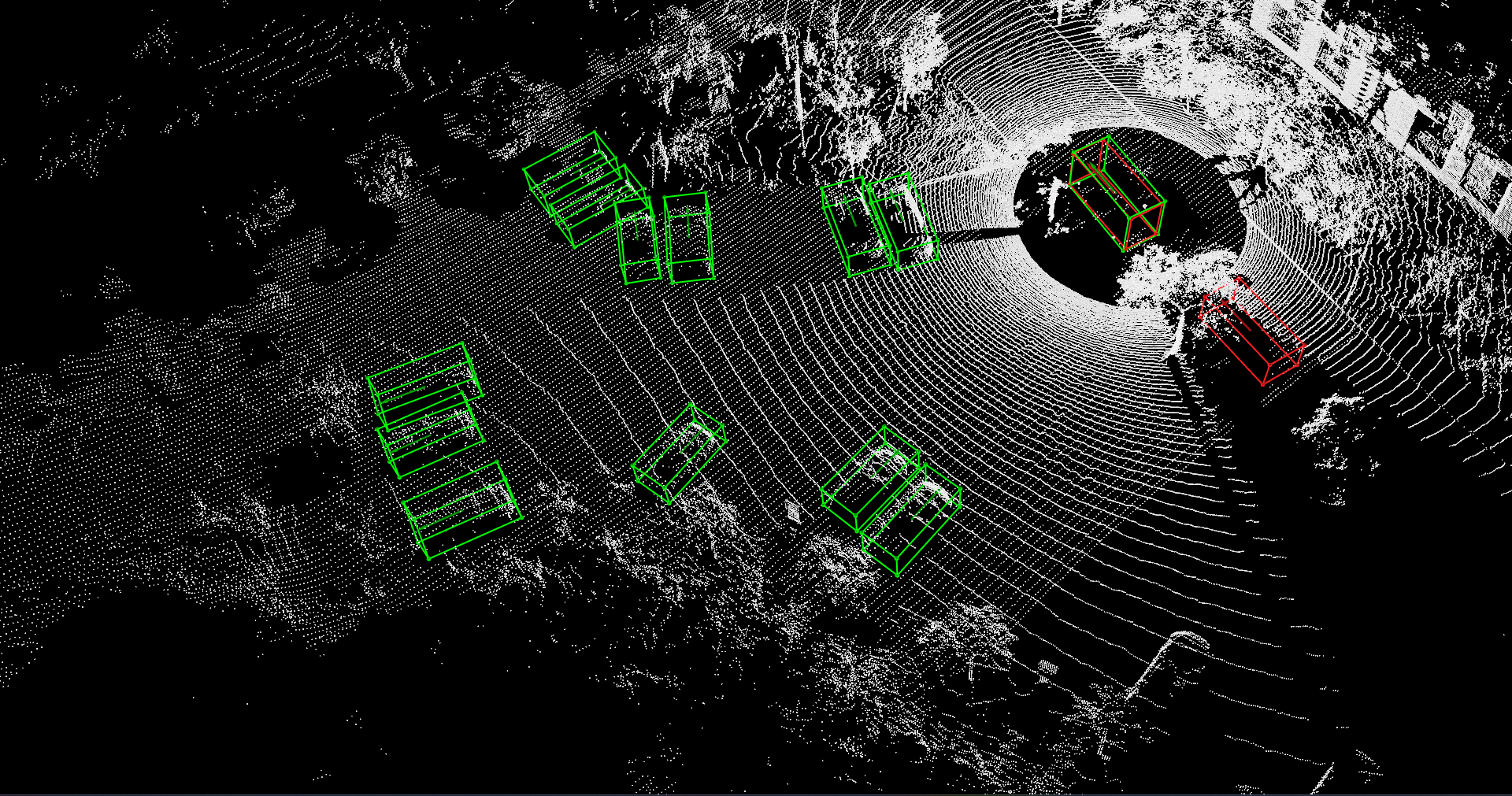} &
\includegraphics[width=0.3\linewidth,height=0.17\linewidth]{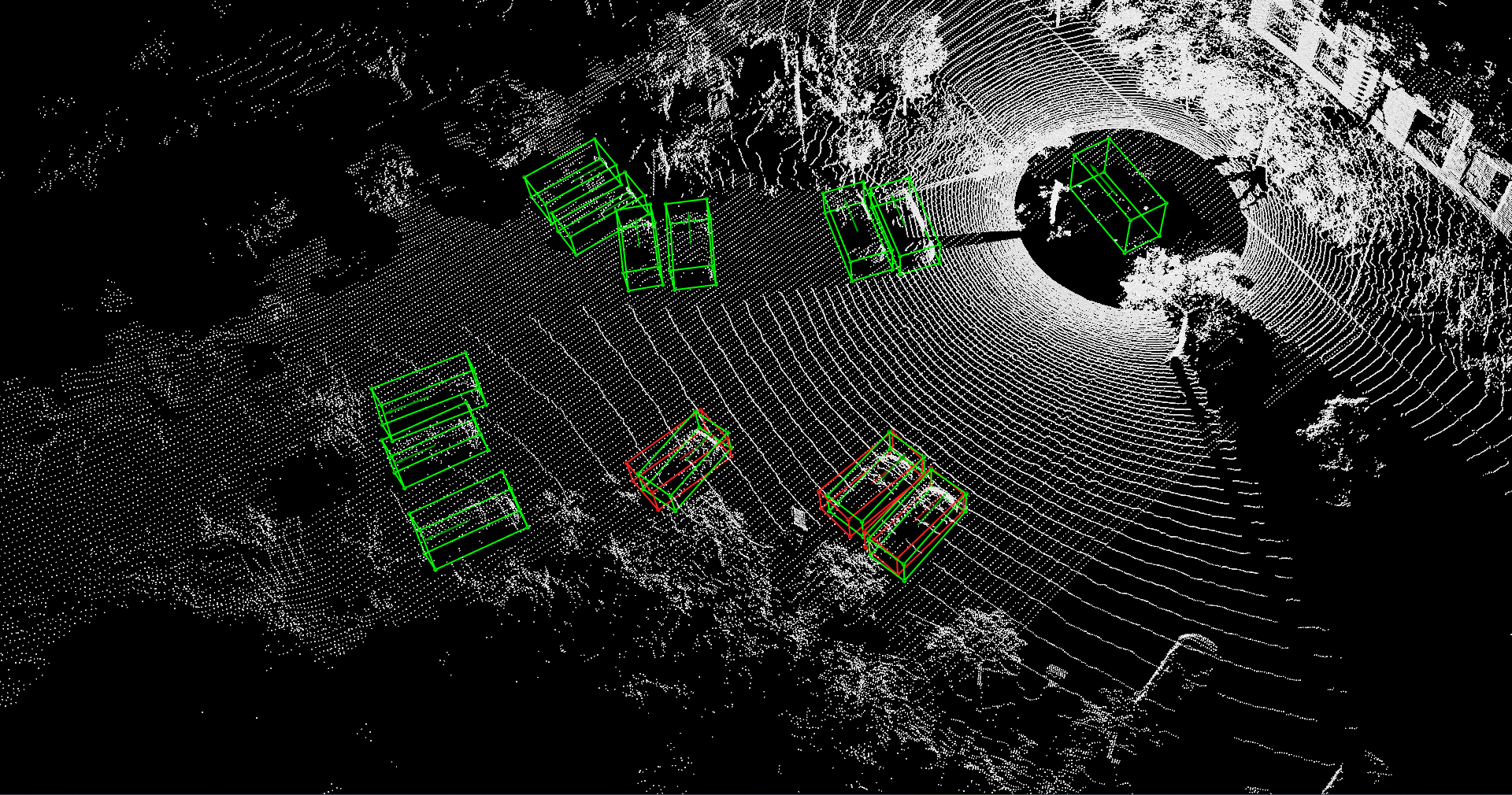} &
\includegraphics[width=0.3\linewidth,height=0.17\linewidth]{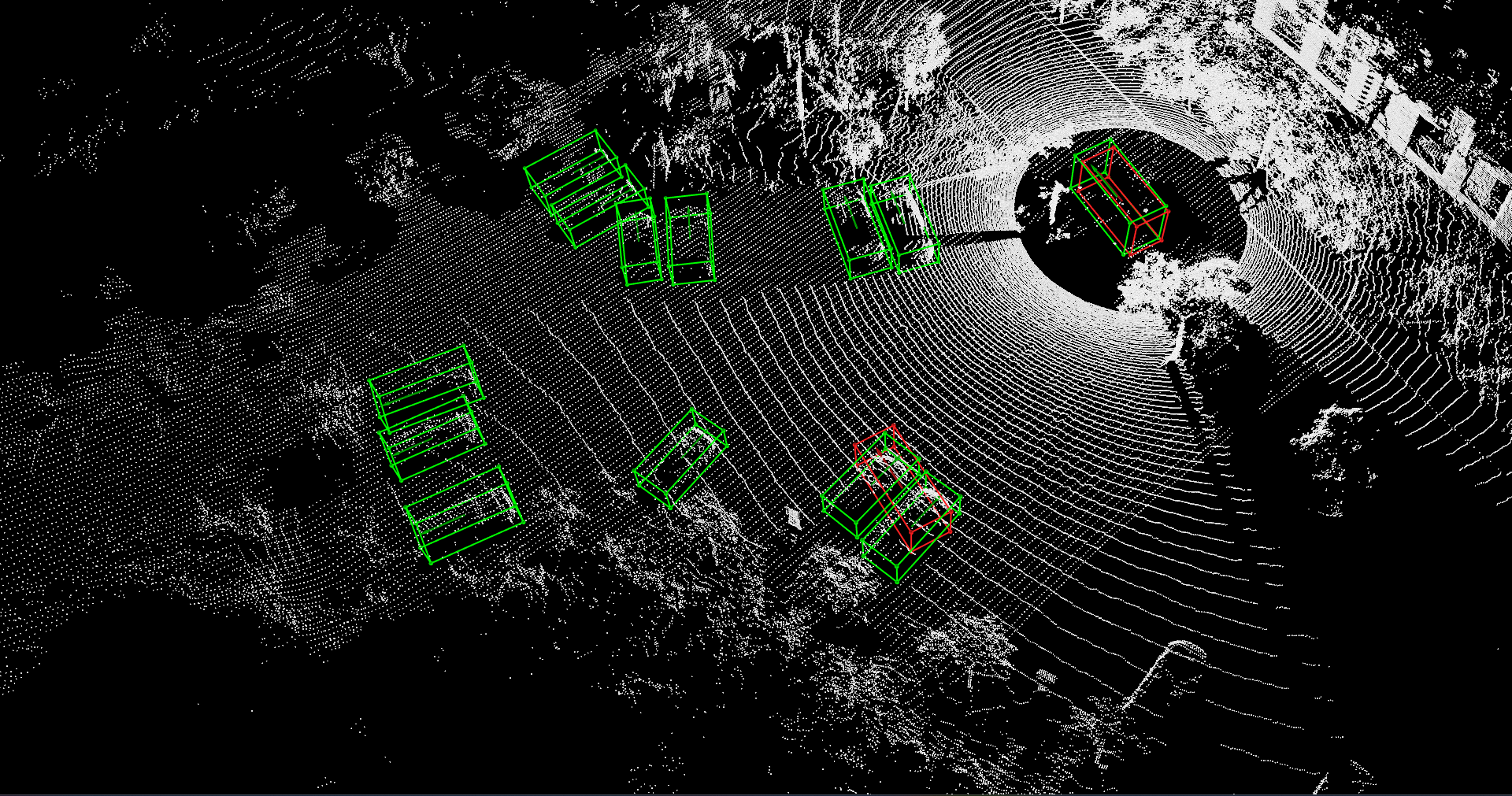} \\
AttFuse~\cite{xu2022opv2v} & CoAlign~\cite{lu2023robust} & DSRC~\cite{zhang2025dsrc} \\
\end{tabular}

\vspace{-3mm}
\caption{\textbf{Visualization of 3D cooperative detection results scene 2.} \textcolor{green}{Green bounding boxes} indicate ground-truth annotations. \textcolor{red}{Red bounding boxes} indicate the model's predictions.}
\label{fig:vis_2}
\vspace{-4mm}
\end{figure*}

\begin{figure*}[!ht]
\centering
\footnotesize
\begin{tabular}{ccc}

\includegraphics[width=0.3\linewidth,height=0.16\linewidth]{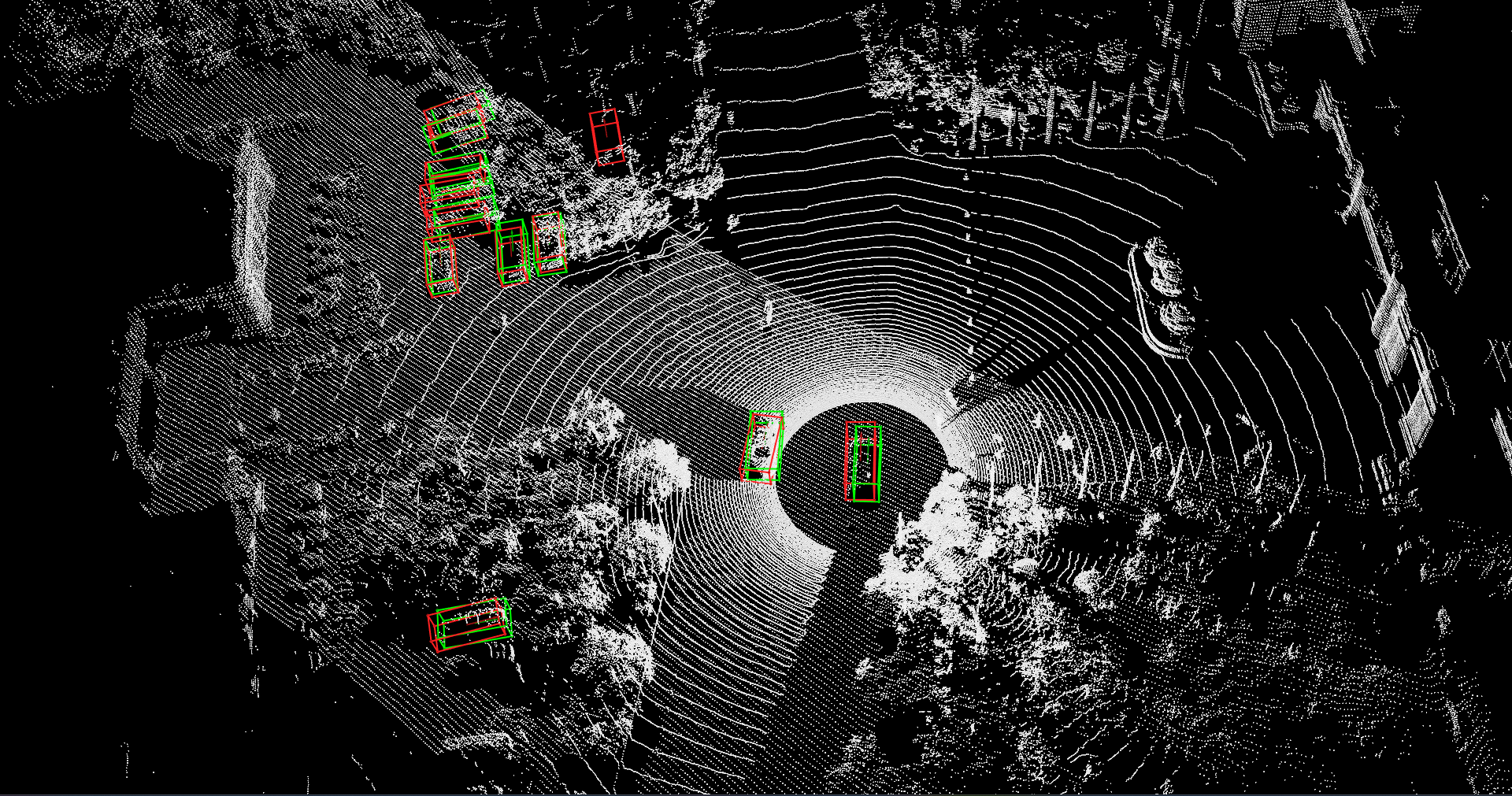} &
\includegraphics[width=0.3\linewidth,height=0.16\linewidth]{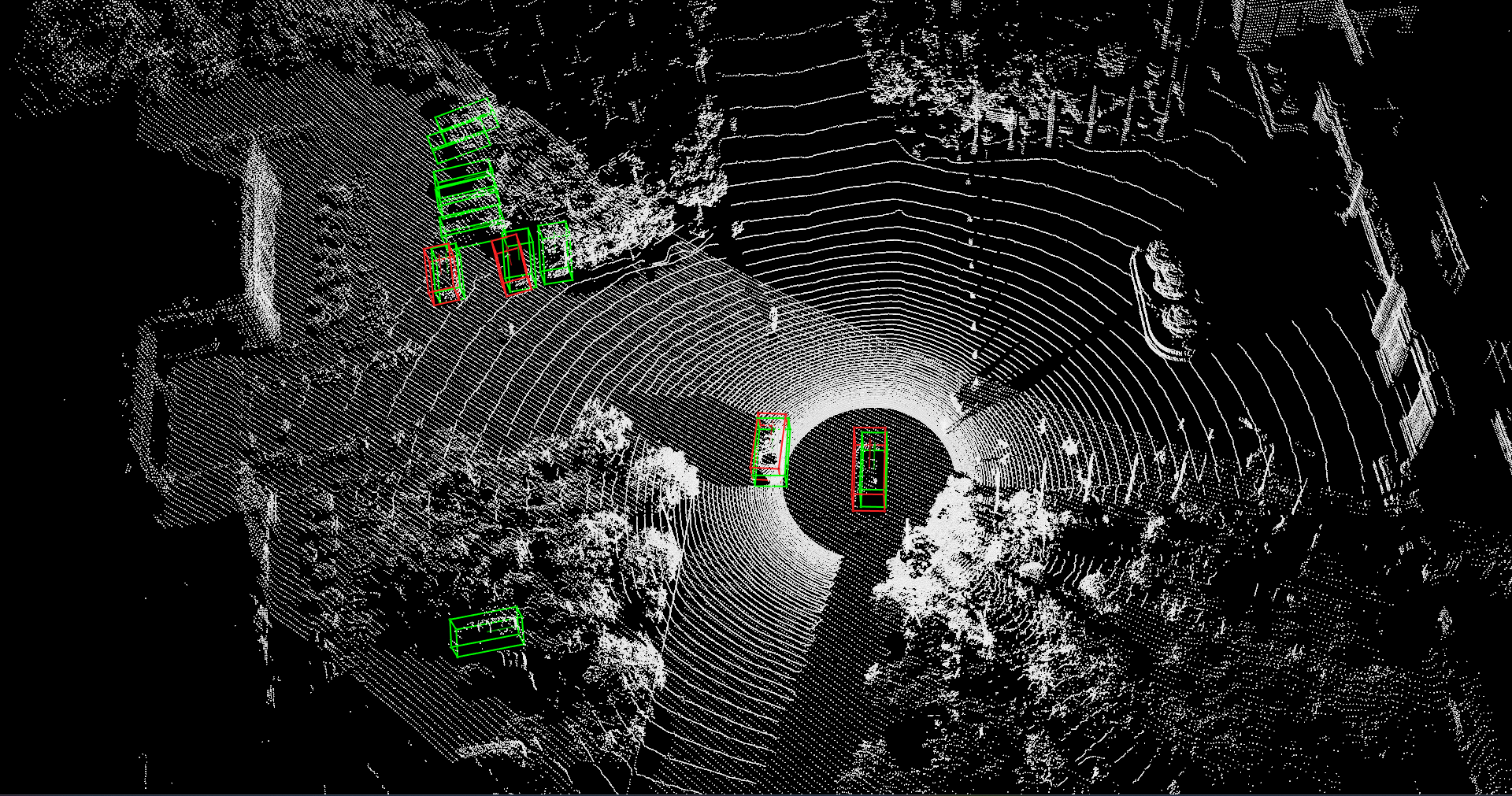} &
\includegraphics[width=0.3\linewidth,height=0.16\linewidth]{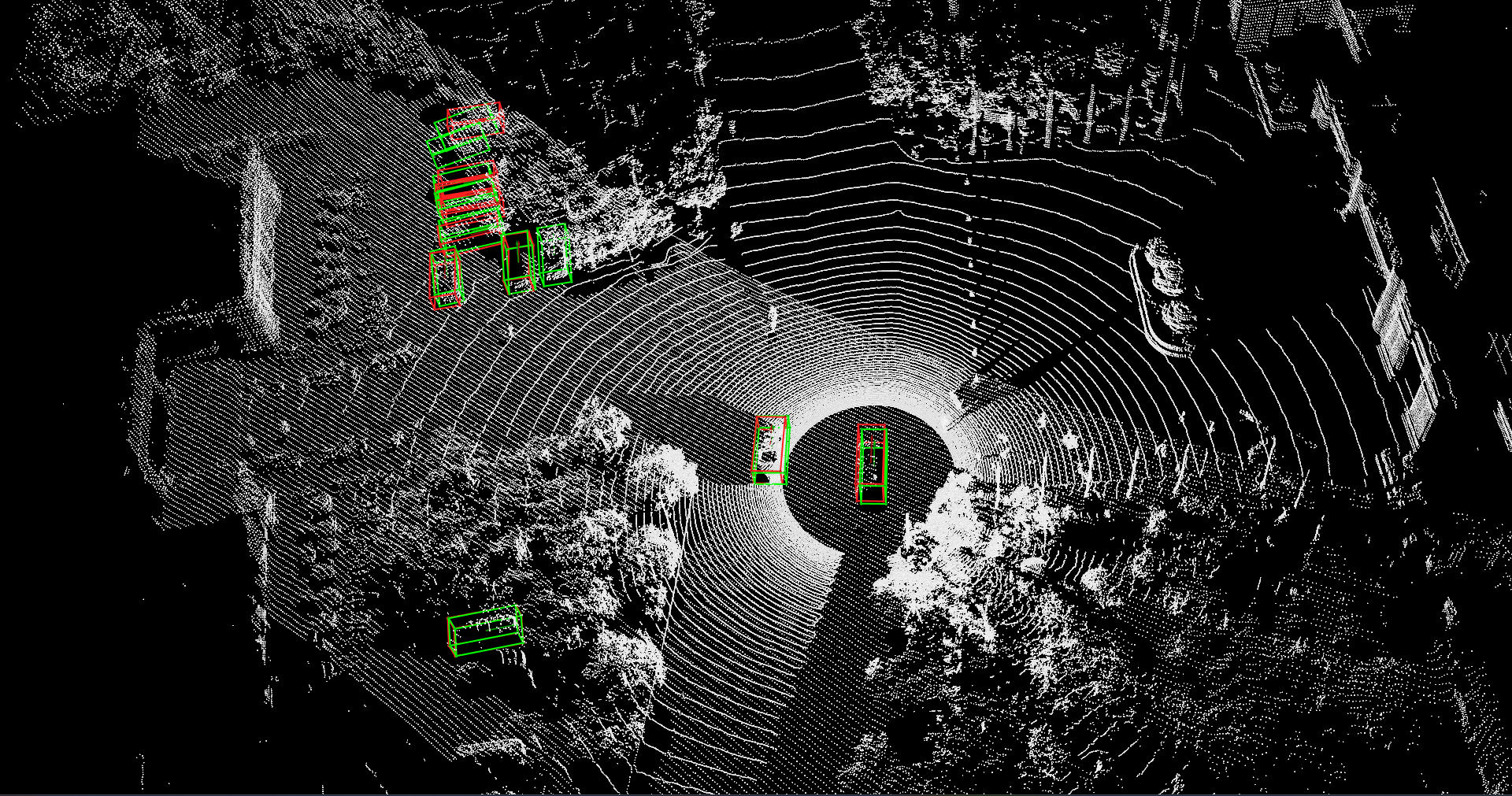} \\
Early Fusion & Late Fusion &  Where2comm~\cite{hu2022where2comm}\\
\includegraphics[width=0.3\linewidth,height=0.17\linewidth]{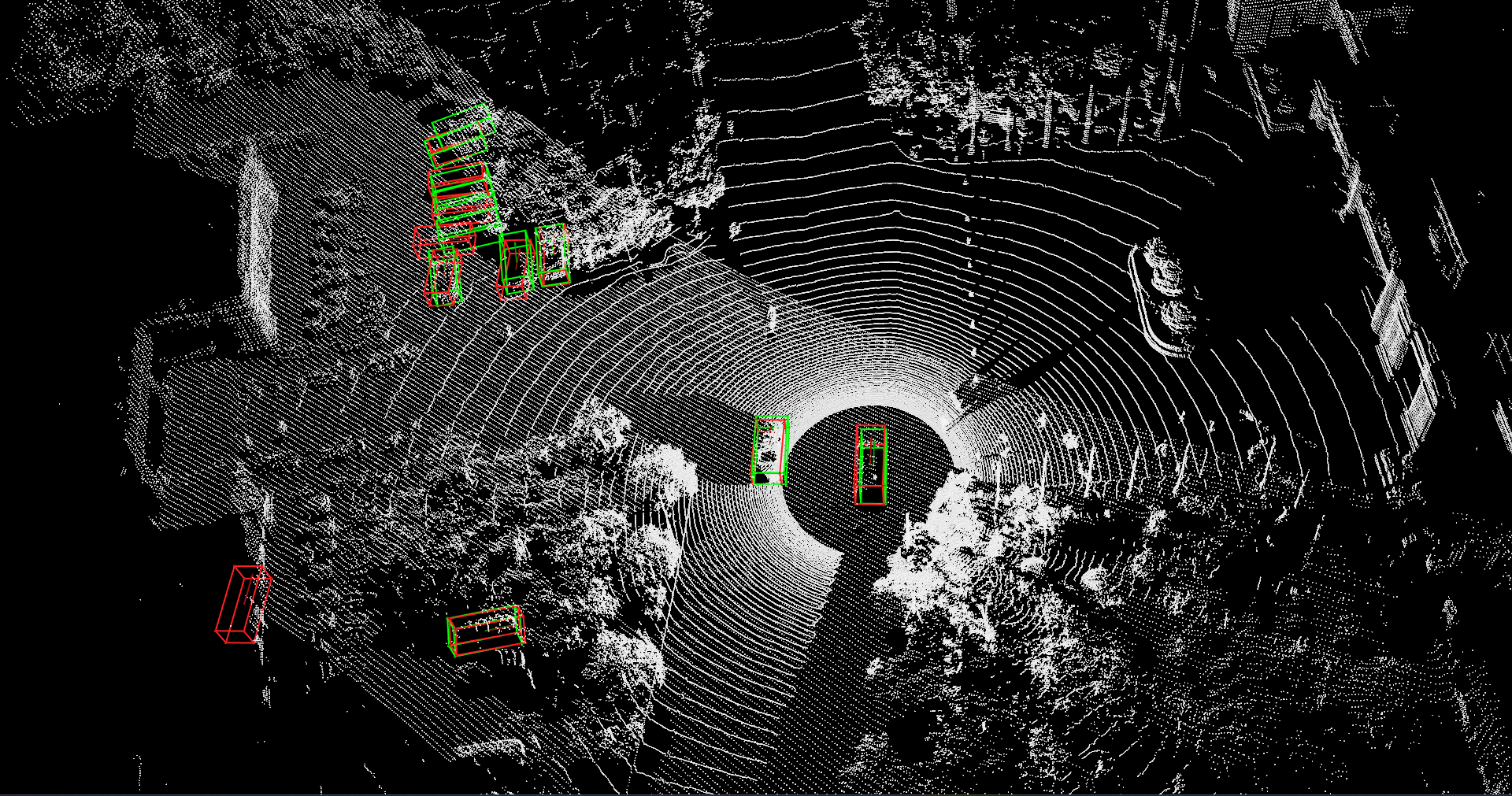} &
\includegraphics[width=0.3\linewidth,height=0.17\linewidth]{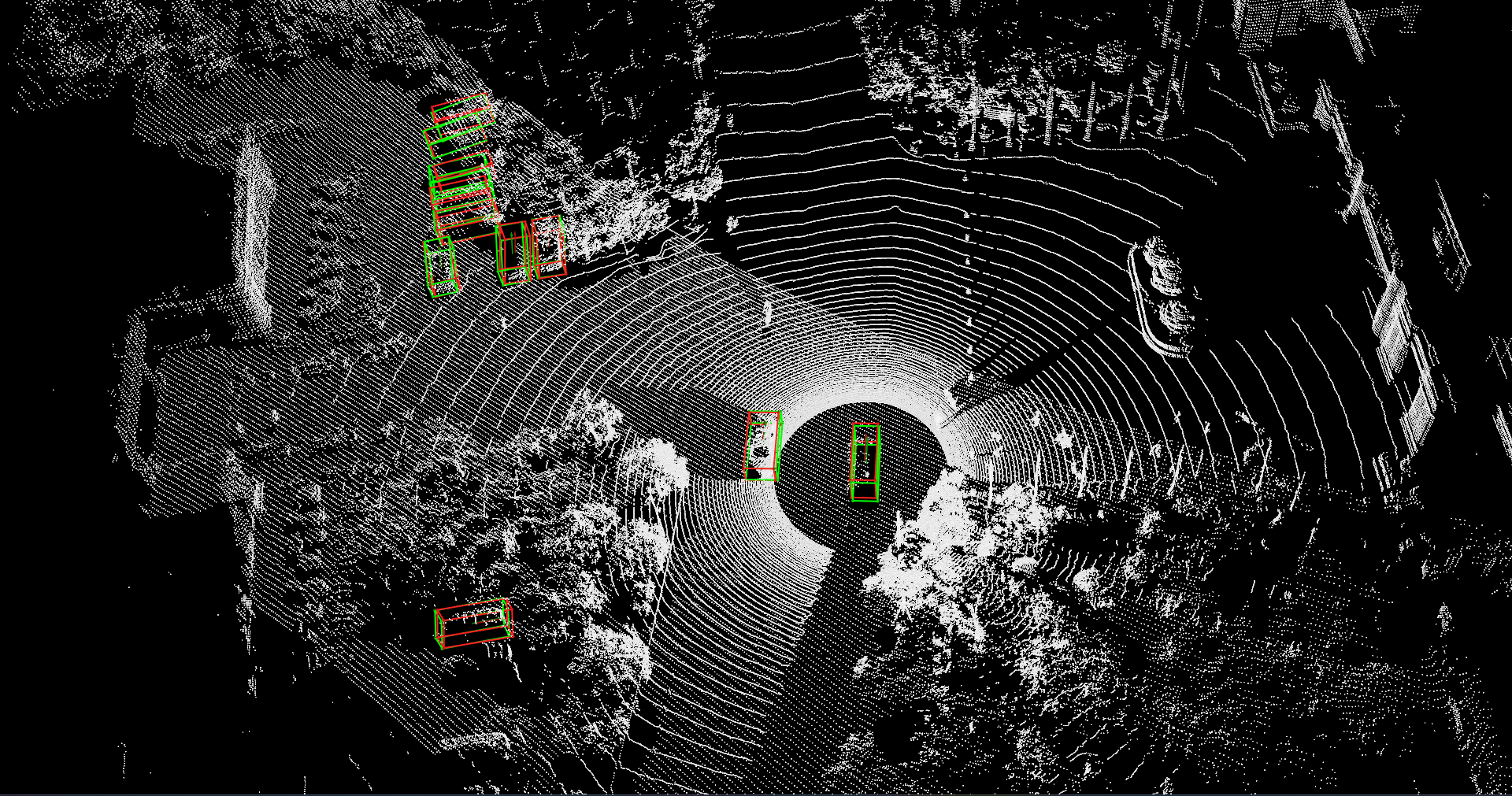} &
\includegraphics[width=0.3\linewidth,height=0.17\linewidth]{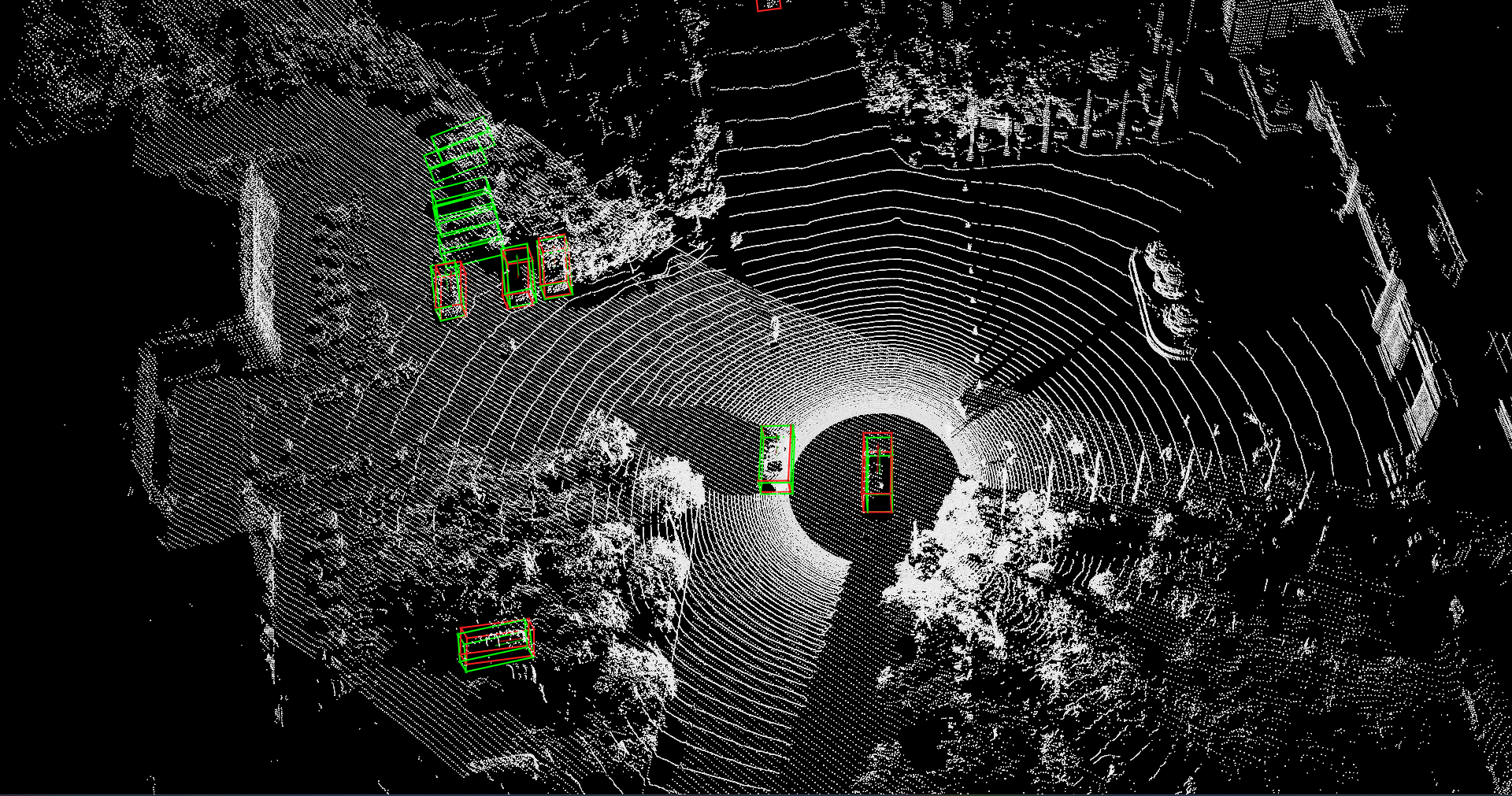} \\
AttFuse~\cite{xu2022opv2v} & CoAlign~\cite{lu2023robust} & DSRC~\cite{zhang2025dsrc} \\
\end{tabular}

\vspace{-3mm}
\caption{\textbf{Visualization of 3D cooperative detection results scene 3.} \textcolor{green}{Green bounding boxes} indicate ground-truth annotations. \textcolor{red}{Red bounding boxes} indicate the model's predictions.}
\label{fig:vis_3}
\vspace{-4mm}
\end{figure*}

We present additional qualitative results of cooperative 3D detection comparisons in Fig.~\ref{fig:vis_1}, \ref{fig:vis_2}, and \ref{fig:vis_3}, all evaluated under the synchronous setting. In these scenes, Intermediate Fusion methods demonstrate significantly better detection performance compared to Late Fusion and Early Fusion. CoAlign~\cite{lu2023robust} achieves the closest alignment between predicted and ground-truth bounding boxes among all compared methods. These observations are consistent with the numerical results reported in Tab.~\ref{tab:exps_coperception} and reflect the internal mechanisms of each algorithm. We further analyze typical failure cases across methods. Specifically, we observe misalignment between \textbf{\textcolor{red}{predicted}} and \textbf{\textcolor{green}{ground-truth}} bounding boxes, which is mainly caused by localization noise arising from different agent motion patterns (Fig.~\ref{fig:motivation} in the main paper). In addition, missed detections are frequently observed under viewpoint inconsistency, where heterogeneous sensing perspectives lead to discrepancies in cross-agent point cloud distributions (Fig.~\ref{fig:Data_analysis}(b) in the main paper).

\begin{table}[t]
\centering
\footnotesize
\caption{\textbf{Implementation details of cooperative 3D object detection task.} CAW stands for CosineAnnealWarm.}
\label{tab:sup_imp_details_cooperative}
\setlength{\tabcolsep}{3pt}
\begin{tabular}{l|c|c|c}
\toprule
Method  & Batch  & LR Scheduler  & LiDAR Range\\
\midrule
No Fusion & 4 &  OneCycle  & $[-100, -80, 100, 80]$\\
Early Fusion  & 4 &  MultiStep & $[-100, -80, 100, 80]$\\
Late Fusion           & 4 &  MultiStep & $[-100, -80, 100, 80]$\\
Where2comm\cite{hu2022where2comm}       & 4  & CAW  & $[-100, -80, 100, 80]$\\
V2X-VIT\cite{xu2022v2x}             & 2  & CAW &  $[-89.6, -76.8, 89.6, 76.8]$\\
AttFuse\cite{xu2022opv2v}           & 4  & MultiStep & $[-100, -80, 100, 80]$\\
CoBEVT\cite{xu2022cobevt}           & 2 & CAW & $[-102.4, -80, 102.4, 80]$\\
CoAlign\cite{lu2023robust}          & 4  & MultiStep & $[-100, -80, 100, 80]$\\
ERMVP\cite{zhang2024ermvp}        & 2  & CAW & $[-89.6, -76.8, 89.6, 76.8]$\\
DSRC\cite{zhang2025dsrc}      & 4 & CAW & $[-100, -80, 100, 80]$\\
\bottomrule
\end{tabular}
\vspace{-4mm}
\end{table}

\subsection{Quantitative Analysis of Localization Noise}
In cooperative perception, the accuracy of relative poses between agents is more critical than absolute pose estimation. Therefore, our evaluation focuses on the consistency of relative transformations rather than absolute pose errors. We leverage the annotated 3D bounding boxes of the ego vehicle and the UAV, which are defined within a shared global coordinate system. Based on these annotations, we derive the deviations of the same target across different agents in each dimension, which indirectly reflect the relative localization errors of GPS. We report the mean ($\mu$) and standard deviation ($\sigma$) of the errors in X, Y, Z, Yaw, and IoU. As shown in Table~\ref{tab:sup_localization_noise}, our dataset achieves comparable or lower noise levels than V2V4Real~\cite{xu2023v2v4real} in most metrics.
\begin{table}[htbp] 
\caption{\textbf{Quantitative results of localization noise.}}
\footnotesize 
\centering 
\renewcommand\arraystretch{0.6}
\setlength{\tabcolsep}{6.2pt}
\begin{tabular}{c|ccccc} 
\toprule 
\multirow{2}{*}{Dataset} & \multicolumn{5}{c}{Localization Noise ($\mu$ / $\sigma$)} \\ \cmidrule(lr){2-6} & X (m) & Y (m) & Z (m) & Yaw ($^\circ$) & IoU (\%)\\ 
\midrule V2V4Real & 7.0/48.4 & 0.6/20.3 & 0.1/1.4 & 0.05/1.5 & 2.4/11.0 \\ Ours & 2.4/6.8 & 1.6/2.9 & 1.0/1.1 & 0.1/1.5 & 4.6/11.6 \\ 
\bottomrule 
\end{tabular} 
\label{tab:sup_localization_noise} 
\vspace{-4mm}
\end{table}

\section{Limitations and Future Work}
\label{sec:limitations}

While our experiments on the V2U4Real dataset demonstrate the promise of 3D V2U cooperative perception, several important limitations remain. The current approach primarily focuses on LiDAR-based V2U cooperation, leaving multi-modal sensing underexplored, and multi-view cooperative strategies have not been fully investigated. The scale and heterogeneity of collaborating agents are also limited, whereas real-world deployments typically involve dynamic networks of diverse vehicles and UAVs with varying sensing and communication capabilities. Moreover, due to airspace management policies and safety regulations, UAV data collection is restricted to designated regions, which may limit the diversity and coverage of real-world scenarios. In future work, we will further expand the dataset to more challenging scenarios, including off-road environments, nighttime conditions, adverse weather, varying flight altitudes, and broader operational ranges.

\def\imgW{0.31\linewidth}
\def\imgH{0.15\linewidth}
\def\im_shift{0\textwidth}

\end{document}